\documentclass[11pt]{article}

\usepackage{acl}

\usepackage{times}
\usepackage{latexsym}
\usepackage[T1]{fontenc}
\usepackage[utf8]{inputenc}
\usepackage{microtype}
\usepackage{inconsolata}

\usepackage{enumitem}
\setlist[itemize]{label=\textbullet, leftmargin=*}
\usepackage{float}
\usepackage{booktabs}
\usepackage{multirow}
\usepackage{tabularx}

\usepackage{amsmath,amssymb,mathtools,amsthm,bm}
\usepackage{graphicx}
\usepackage{subcaption}
\usepackage[table]{xcolor}
\usepackage[most]{tcolorbox}

\newtcolorbox{cadre}[1][]{enhanced, breakable,
  colback=gray!5, colframe=black!40, boxrule=0.5pt,
  arc=2mm, left=1em, right=1em, top=0.6em, bottom=0.6em,
  fonttitle=\bfseries, title=Cadre, #1}

\theoremstyle{definition}
\newtheorem{definition}{Definition}[section]

\theoremstyle{plain}

\usepackage{algorithm}
\usepackage{algpseudocode}

\usepackage{tikz}
\usetikzlibrary{arrows.meta, positioning, fit, backgrounds, shapes.geometric, calc}
\usepackage{pgfgantt}

\tikzset{
    >=Latex,
    var/.style={
        circle, draw=black!90, thick, 
        fill=white, minimum size=1.1cm, 
        font=\large\bfseries, align=center
    },
    var_latent/.style={
        circle, draw=blue!70, thick, dashed,
        fill=blue!5, minimum size=1.1cm, 
        font=\large\bfseries, align=center
    },
    latent/.style={
        circle, draw=gray!80, thick, dashed, 
        fill=gray!10, minimum size=1.1cm,
        font=\large\bfseries, align=center
    },
    selection/.style={
        shape=diamond, draw=orange!80!black, thick, dashed,
        fill=orange!5, minimum size=1.2cm,
        font=\large\bfseries, align=center
    },
    arrow/.style={->, thick, draw=black!80},
    arrow_latent/.style={->, thick, dashed, draw=blue!70},
    observed_box/.style={
        draw=red!80, solid, thick, rounded corners=2mm, 
        inner sep=14pt
    }
}

\definecolor{causalgreen}{HTML}{1B5E20}
\definecolor{spuriousblue}{HTML}{0D47A1}
\definecolor{irmorange}{HTML}{E65100}
\definecolor{bgbox}{HTML}{F8F9FA}
\definecolor{bordergray}{HTML}{D1D5DB}

\newcommand{\badgeC}{\colorbox{red!10}{\textcolor{red!80!black}{\textbf{\strut C}}}}
\newcommand{\badgeE}{\colorbox{causalgreen!10}{\textcolor{causalgreen}{\textbf{\strut E}}}}
\newcommand{\ermbubble}[1]{%
  \tikz[baseline=(char.base)]{%
    \node[rectangle, draw=spuriousblue, fill=spuriousblue!10, rounded corners=2pt, inner sep=1pt, outer sep=0pt, minimum size=0pt, thin, align=center, font=\bfseries\tiny] (char) {#1};%
  }%
}

\newcommand{\ms}[2]{%
    \ensuremath{#1 \pm #2}%
}
\newcommand{\professionpair}[2]{#1/\allowbreak#2}

\renewcommand{\ms}[2]{%
  \ensuremath{#1{\scriptstyle\,\pm\,#2}}%
}

\title{Disentangling Spurious Correlations: How IRM Handles Diverse Shortcuts \\ in Text Classification}

\author{
  \textbf{Paul Grunenwald}$^{1,2}$\thanks{\raggedright Contact: \texttt{paul.grunenwald@etu.u-paris.fr}} \quad 
  \textbf{Benoît Crabbé}$^2$ \quad 
  \textbf{Guillaume Wisniewski}$^2$ \quad 
  \textbf{Pirmin Lemberger}$^1$ \\
  $^1$onepoint, Paris, France \\
  $^2$Laboratoire de Linguistique Formelle, Université Paris Cité, CNRS, Paris, France
}

\begin{document}
\maketitle

\begin{abstract}
In NLP, models trained via Empirical Risk Minimization (ERM) often exploit spurious correlations, such as dialectal markers, at the expense of robust linguistic features, thereby degrading their out-of-distribution (OOD) generalization. Invariant Risk Minimization (IRM) aims to suppress such shortcuts by identifying predictive relationships that remain stable across environments, but its practical utility for NLP remains debated. We use a causality-driven framework to evaluate IRM under three shortcut-generating mechanisms: label-generated features, confounding, and selection bias. Our results show that IRM successfully rejects shortcuts across the evaluated causal and anti-causal structures, including under limited environmental variation and increasing label noise. We then extend our analysis beyond controlled synthetic artifacts to two less constrained settings: gender-profession associations in Bias in Bios and semantically loaded actor-name shortcuts in ActorCorr. These experiments show that invariant learning can mitigate more realistic textual shortcuts, while semantically meaningful features require careful calibration of the invariance penalty.
\end{abstract}

\section{Introduction}

Predictive success in NLP often masks a vulnerability: models frequently rely on dataset-specific biases rather than genuine robust linguistic features. Under the ERM paradigm \citep{vapnik2013nature}, models exploit statistical shortcuts, such as negation words \citep{gururangan-etal-2018-annotation}, syntactic heuristics \citep{mccoy-etal-2019-right}, or demographic dialectal markers \citep{blodgett-etal-2020-language}, leading to catastrophic performance drops when encountering real-world distribution shifts where these biases no longer hold. \begin{figure}[t]
    \centering
    \resizebox{\columnwidth}{!}{%
    \begin{tikzpicture}[
        font=\scriptsize,
        envbox/.style={
            rectangle, draw=gray!40, fill=gray!5, rounded corners=3pt, thick,
            inner xsep=2mm, inner ysep=2mm
        },
        textcard/.style={
            rectangle, draw=gray!25, fill=white, rounded corners=2pt,
            text width=0.44\columnwidth, align=left,
            inner xsep=3pt, inner ysep=3pt,
            anchor=north west
        },
        stratbox/.style={
            rectangle, draw=#1, fill=#1!4, rounded corners=3pt, thick,
            text width=0.93\columnwidth,
            inner xsep=6pt, inner ysep=6pt,
            anchor=north, align=left
        }
    ]

        
        \node[anchor=north west, font=\bfseries\scriptsize, text=gray!80!black, inner sep=0pt] (title1) at (0,0) {Environment $e_1$};
        \node[anchor=north west, font=\bfseries\scriptsize, text=gray!80!black, inner sep=0pt] (title2) at (0.51\columnwidth,0) {Environment $e_2$};
        
        \node[textcard] (ex11) at ([yshift=-2mm]title1.south west) {\textbf{1. P:} He is sleeping. \\ \textbf{H:} He is \ermbubble{not} awake. \hfill \badgeC};
        \node[textcard] (ex12) at ([yshift=-1.5mm]ex11.south west) {\textbf{2. P:} The shop is closed. \\ \textbf{H:} It is \ermbubble{not} open. \hfill \badgeC};
        \node[textcard] (ex13) at ([yshift=-1.5mm]ex12.south west) {\textbf{3. P:} The box is empty. \\ \textbf{H:} It is \ermbubble{not} full. \hfill \badgeC};
        \node[textcard] (ex14) at ([yshift=-1.5mm]ex13.south west) {\textbf{4. P:} A child is jumping. \\ \textbf{H:} A person is moving. \hfill \badgeE};

        \node[textcard] (ex21) at ([yshift=-2mm]title2.south west) {\textbf{1. P:} The worker is tired. \\ \textbf{H:} He is \ermbubble{not} rested. \hfill \badgeC};
        \node[textcard] (ex22) at ([yshift=-1.5mm]ex21.south west) {\textbf{2. P:} The room is dark. \\ \textbf{H:} The room is bright. \hfill \badgeC};
        \node[textcard] (ex23) at ([yshift=-1.5mm]ex22.south west) {\textbf{3. P:} The window is open. \\ \textbf{H:} It is \ermbubble{not} closed. \hfill \badgeE};
        \node[textcard] (ex24) at ([yshift=-1.5mm]ex23.south west) {\textbf{4. P:} She is reading. \\ \textbf{H:} She is studying. \hfill \badgeE};

        \begin{scope}[on background layer]
            \node[envbox, fit=(title1) (ex14)] (e1) {};
            \node[envbox, fit=(title2) (ex24)] (e2) {};
        \end{scope}

        \coordinate (midbottom) at ($(e1.south east)!0.5!(e2.south west)$);

        \node[stratbox=spuriousblue] (erm) at ([yshift=-1mm]midbottom) {
            \begin{tabularx}{\linewidth}{@{}lX@{}}
                \textbf{\textcolor{spuriousblue}{ERM}} & Concatenates $e_1 \cup e_2$. Observes that \ermbubble{not} is present in 4/5 of \badgeC\ and absent in 2/3 of \badgeE. \\[1mm]
                & $\implies$ \textbf{\textcolor{spuriousblue}{Learns shortcut:}} \ermbubble{not} $\to$ \badgeC
            \end{tabularx}
        };
    
        \node[stratbox=causalgreen] (irm) at ([yshift=-1mm]erm.south) {
            \begin{tabularx}{\linewidth}{@{}lX@{}}
                \textbf{\textcolor{causalgreen}{IRM}} & Distinguishes $e_1, e_2$. Observes the \ermbubble{not} $\to$ \badgeC\ rule works in $e_1$ (3/3) but drops to 50\% (1/2) in $e_2$. \\[6mm]
                & $\implies$ \textbf{\textcolor{causalgreen}{Rejects unstable shortcut}}.
            \end{tabularx}
        };
    \end{tikzpicture}%
    }
    
    \caption{\textbf{Optimization dynamics under multi-environment training.} On a binary classification task, ERM exploits a global spurious correlation, while IRM leverages environment boundaries to detect its instability and discard it. (C: Contradiction, E: Entailment).}
    \label{fig:erm_vs_irm}
\end{figure}
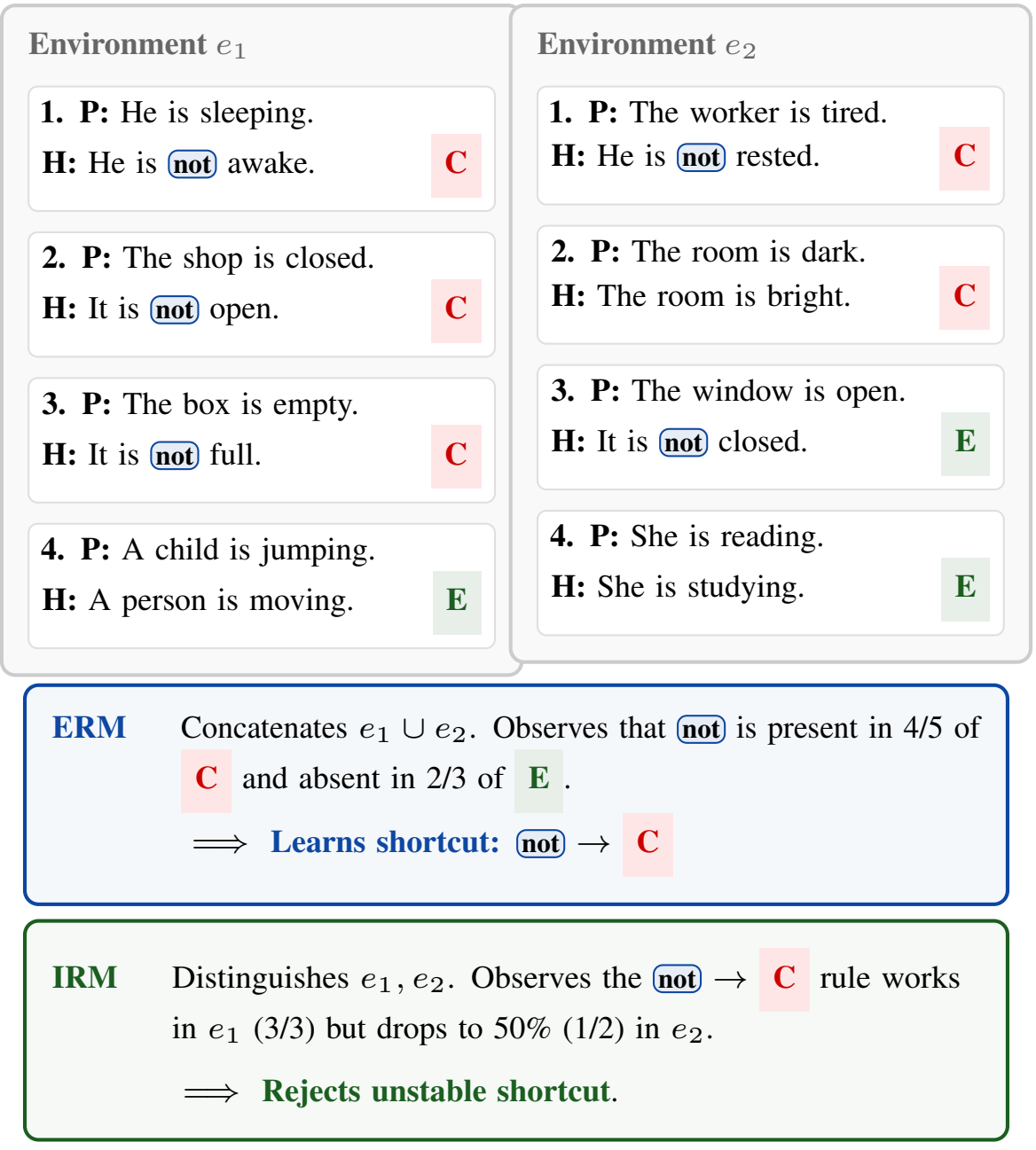

To mitigate this lack of robustness, several strategies have been proposed. Domain adaptation methods \citep{ben-david2010theory, ganin2016domain} and sample reweighting techniques \citep{shimodaira2000improving} aim to align distributions or correct known selection biases. However, these techniques face a major limitation: they generally require prior knowledge of the test distribution, either through access to unlabeled target data or by explicitly identifying the specific confounding variables to be rebalanced. In many real-world NLP applications, this target distribution is unknown or subject to unpredictable shifts. In this context, IRM \citep{arjovsky2019invariant} has emerged as a powerful theoretical alternative. The core originality of IRM lies in its ability to operate without any information regarding the target distribution. By leveraging data partitioned into multiple distinct training \emph{environments}, IRM enforces a strict learning constraint. It compels the representation to discard features whose predictive validity fluctuates, isolating only those stable, invariant correlations that yield simultaneously optimal classifiers across all environments. In doing so, IRM seeks to capture robust linguistic principles that guarantee optimal generalization, even to domains entirely unseen during training (Figure~\ref{fig:erm_vs_irm}). Despite these theoretical promises, the practical effectiveness of IRM remains heavily debated. While large-scale benchmarks in machine learning have questioned its out-of-the-box superiority over ERM in standard domain generalization \citep{gulrajani2020search}, evaluations on specific NLP tasks have also tempered expectations \citep{koh2021wilds}, notably regarding its capacity for robust bias mitigation \citep{dranker2021irm}. To understand why cross-environment invariant methods lack effectiveness in NLP, we must move beyond treating spurious correlations as an undifferentiated problem. A non-causal correlation between a spurious textual variable and the label $Y$ can originate from distinctly different mechanisms: a common cause, a selection bias, or a label-generated feature \citep{pearl2009causality}. As these distinct generative processes result in structurally different distributional shifts, investigating how IRM responds to each specific spurious mechanism might help clarify the precise conditions required for its success in NLP. Furthermore, IRM's core assumption, that spurious correlations must fluctuate across environments, raises questions about the required magnitude of variation and its behavior when the genuine task-relevant signal is degraded.

\paragraph{Our contributions.} We provide a causal analysis of when IRM mitigates textual shortcuts. First, Section~\ref{sec:environment-construction} distinguishes three shortcut-generating mechanisms, label-generated features, confounding, and selection bias, across causal and anti-causal tasks. Second, Sections~\ref{sec:sensitivity-analysis} and \ref{sec:noise-sweep} quantify how success depends on cross-environment variation and task-signal reliability. Third, Section~\ref{sec:beyond-synthetic} extends the analysis to naturally occurring gender-profession correlations and semantically loaded actor-name shortcuts, revealing an accuracy-invariance trade-off.

\section{Background}

\subsection{Problem Setting: OOD Generalization}

At the core of current challenges in NLP lies the issue of robustness to change, as models are rarely deployed under linguistic conditions identical to those encountered during training. To formalize this heterogeneity, we adopt the concept of \emph{environment} from the invariant learning literature \citep{peters2016causal, arjovsky2019invariant}. Intuitively, an environment corresponds to a distinct data source, such as formal book reviews versus informal tweets, where the underlying task is identical, but the specific vocabulary, style, or local biases differ. Each environment $e$ represents a specific generative context (such as a particular domain or demographic group) characterized by a joint probability distribution $P^e(X,Y)$ over texts $X$ and labels $Y$. We denote by $\mathcal{E}_{\text{all}}$ the comprehensive set of all possible environments across which we evaluate a model's ability to achieve OOD generalization. The goal is to find $f$ that minimizes the risk $R^e(f)$ in the worst-case scenario across all possible environments:
\[
\min_f \max_{e \in \mathcal{E}_{\text{all}}} R^e(f),
\]
where $R^e(f) = \mathbb{E}_{(x,y)\sim P^e}[\ell(f(x),y)]$ represents the expected risk under environment $e$ for a given loss function $\ell$.

\subsection{Empirical Risk Minimization and Its Limitations}

The standard paradigm for predictive learning is ERM, which relies on a strict stationarity assumption: it presumes that the training data is independent and identically distributed (i.i.d.). It assumes that the global training distribution $P_{\text{train}}$, which is defined as the mixture of the joint distributions from all training environments $e \in \mathcal{E}_{\text{tr}}$, is strictly identical to the test distribution $P_{\text{test}}$ (representing a target environment $e_{\text{test}} \notin \mathcal{E}_{\text{tr}}$). In practice, given a subset of training environments $\mathcal{E}_{\text{tr}} \subset \mathcal{E}_{\text{all}}$, ERM minimizes the average empirical risk $R_{\text{ERM}}(f)$ over a pooled dataset of size $n$:
\begin{equation}
\hat{f}_{\text{ERM}} = \arg\min_f \frac{1}{n} \sum_{i=1}^n \ell(f(x_i), y_i)
\label{eq:erm}
\end{equation}
By blindly minimizing this pooled loss, the model exploits all statistical correlations globally present in $P_{\text{train}}$. This is the core vulnerability of ERM: it fails to distinguish between robust features and spurious correlations. In an OOD scenario where $P_{\text{train}} \neq P_{\text{test}}$, these shortcuts vanish, inevitably causing the model's performance to collapse.

\subsection{Invariant Risk Minimization}
\label{sec:irm}

To address this generalization failure, IRM leverages the distributional diversity across the available training environments $\mathcal{E}_{\text{tr}}$. Within this multi-environment framework, the fundamental intuition is to extract a unique feature representation $\Phi$ such that a single classifier $w$ built on top of it remains simultaneously optimal across all contexts. The explicit role of $\Phi$ is to capture exclusively the stable correlations that persist across environments. By enforcing this strict invariance constraint, the model is compelled to discard environment-specific spurious correlations and rely solely on these robust features. Formally, IRM is framed as a bi-level optimization problem:
\begin{equation}
\begin{aligned}
\min_{\Phi, w} \quad & \sum_{e \in \mathcal{E}_{\text{tr}}} R^e(w \circ \Phi) \\
\text{s.t.} \quad & w \in \arg\min_{\tilde{w}} R^e(\tilde{w} \circ \Phi), \forall e \in \mathcal{E}_{\mathrm{tr}}.
\end{aligned}
\label{eq:irm-bilevel}
\end{equation}

Because this is an intractable bilevel optimization problem, a practical relaxation known as IRMv1 is typically used \citep{arjovsky2019invariant, choe2020empirical, dranker2021irm}. In IRMv1, \(\Phi\) is parameterized to directly output predictive logits, while the classifier \(w\) is reduced to a scalar dummy variable fixed at \(w=1\). Invariance is encouraged by penalizing the gradient of each environment-specific risk with respect to this dummy classifier:
\begin{equation}
\label{eq:irmv1}
\min_{\Phi} \sum_{e\in\mathcal E_{\mathrm{tr}}}
\left[
R^e(\Phi)
+ \lambda \left\| \nabla_{w\mid w=1} R^e(w\cdot\Phi) \right\|^2
\right].
\end{equation}
The dummy classifier \(w\) is not a trainable model component. Gradients of the penalty are propagated through \(\Phi\), whose parameters are optimized. The multiplier $\lambda$ acts as a hyperparameter controlling the trade-off between standard empirical risk minimization and the invariance of the representation (enforced via the gradient penalty). We provide a sensitivity analysis of $\lambda$ in Appendix~\ref{app:model-setup}, where we examine its impact on model performance. Crucially, the success of IRM relies on the assumption that spurious correlations vary across $\mathcal{E}_{\text{tr}}$, while task-relevant predictive relationships remain invariant. This dependency motivates our environment construction strategy in Section~\ref{sec:experiment-design}, where we deliberately generate training sets by varying the strength of spurious correlations across environments, thereby creating the necessary heterogeneity to rigorously evaluate IRM's invariance.

\section{Related Work}
\label{sec:related-work}

\paragraph{Applications of invariant learning in NLP.} Applications of invariant learning to NLP have produced mixed results. In the WILDS benchmark, \citet{koh2021wilds} evaluated IRM alongside several other methods for improving robustness to naturally occurring domain and subpopulation shifts, including NLP tasks, but found no consistent advantage over standard training. Focusing more directly on natural language inference, \citet{dranker2021irm} similarly reported limited benefits when natural MNLI genres were treated as environments. However, IRM performed more favorably when environments were explicitly constructed to differ in the prevalence of a known bias, emphasizing the importance of environment definition. Other studies have adapted invariant learning more directly to NLP-specific objectives. \citet{peyrard2022invariant} introduced an invariant language-modeling objective and obtained clearer benefits in controlled bias-removal experiments than across broad, heterogeneous corpora. For demographic bias mitigation, \citet{zhou2023causal} proposed Causal-Debias, which identifies bias-relevant factors as non-causal and label-relevant factors as causal, intervenes on the former across demographic groups, and applies an invariant learning loss during downstream fine-tuning. Their experiments show that this task-aware formulation can reduce stereotypical associations while preserving downstream performance. Taken together, these studies suggest that invariant learning is most effective when the environments or interventions are explicitly designed to expose the targeted bias. Our work complements these approaches by studying whether IRM's behavior depends on the mechanism that generates the spurious correlation and by extending the evaluation from controlled synthetic shortcuts to naturally occurring and semantically loaded signals.

\paragraph{Alternative approaches to shortcut mitigation.} Other mitigation methods rely on different signals to identify shortcut-conflicting examples. Group DRO minimizes the worst-group risk but requires annotations for the relevant subpopulations \citep{sagawa2019distributionally}. JTT removes this requirement by upweighting examples misclassified by an initial ERM model \citep{liu2021just}, whereas LfF uses a deliberately biased model and assumes that spurious features are easier to learn than task-relevant ones \citep{nam2020learning}. Bias-model approaches similarly reweight examples according to their compatibility with a known superficial heuristic \citep{utama2020towards}. IRM uses a distinct source of supervision: rather than identifying biased examples directly, it suppresses predictive relationships that fail to support a common classifier across environments. It therefore does not require the shortcut to be explicitly annotated or easier to learn, but it does require the environment partition to expose variation in its association with the target.

\section{Experiment Design}
\label{sec:experiment-design}

\subsection{Model Architecture}
\label{sec:model-architecture}

All experiments share the same two-component architecture: a frozen DistilBERT encoder \citep{sanh2019distilbert} $\Phi_{\text{BERT}}: \mathcal{X} \rightarrow \mathbb{R}^{768}$, which extracts fixed sentence-level embeddings via masked mean pooling, followed by a single trainable linear head $W_\theta \in \mathbb{R}^{d_{\text{out}} \times 768}$, where $d_{\text{out}}=1$ for binary classification and $d_{\text{out}}=C$ for $C$-class classification. To connect this architecture with the IRMv1 notation introduced in Section~\ref{sec:irm}, we define the complete logit-producing map as $\Phi_\theta(x) = W_\theta \Phi_{\text{BERT}}(x)$. Thus, $\Phi_\theta$ plays the role of the trainable classifier $\Phi$ in the IRMv1 objective, even though the pretrained encoder $\Phi_{\text{BERT}}$ itself remains frozen. Its trainable parameters $\theta$ correspond exclusively to the linear head $W_\theta$. ERM and IRMv1 consequently optimize exactly the same model parameters $W_\theta$ and differ only in their training objective. ERM minimizes the standard classification loss over the pooled training examples, whereas IRMv1 retains the two training environments $\{e_1,e_2\}$ and optimizes the sum of the environment-specific risks augmented by the invariance penalty from Equation~\ref{eq:irmv1}.

Keeping $\Phi_{\text{BERT}}$ frozen is a deliberate design choice. Because $W_\theta$ constitutes the only trainable component under both ERM and IRMv1, the two methods are compared within the same fixed pretrained representation space and with an identical set of trainable parameters. This places our analysis in a controlled linear setting, in which any behavioral difference between ERM and IRMv1 can be attributed to the optimization objective rather than to encoder fine-tuning dynamics or differences in model capacity. This design therefore isolates the effect of the invariance objective, which is the primary object of study.

\subsection{Environment Construction}
\label{sec:environment-construction}

\begin{figure}[!htbp]
    \centering
    \resizebox{0.7\linewidth}{!}{
    \begin{tikzpicture}
        \tikzset{
            leg_var/.style={circle, draw=black!90, line width=1.5pt, fill=white, minimum size=1.2cm},
            leg_var_latent/.style={circle, draw=blue!70, line width=1.5pt, dashed, fill=blue!5, minimum size=1.2cm}
        }
        \node[leg_var] (l1) at (0,0) {};
        \node[right=0.5cm of l1, font=\Huge\bfseries, text=black!90] {Observed Label};
        \node[leg_var_latent] (l2) at (10,0) {}; 
        \node[right=0.5cm of l2, font=\Huge\bfseries, text=blue!80!black] {Text Components};
    \end{tikzpicture}
    }
    \resizebox{0.75\linewidth}{!}{
    \begin{tikzpicture}
        \tikzset{
            leg_latent/.style={circle, draw=gray!80, line width=1.5pt, dashed, fill=gray!10, minimum size=1.2cm},
            leg_env/.style={rectangle, draw=black!80, line width=1.5pt, fill=black!10, minimum size=1.0cm}
        }
        \node[leg_latent] (l3) at (0,0) {};
        \node[right=0.5cm of l3, font=\Huge\bfseries, text=gray!80!black] {Unobserved Concept};
        \node[leg_env] (l4) at (12,0) {}; 
        \node[right=0.5cm of l4, font=\Huge\bfseries, text=black!80] {Environment};
    \end{tikzpicture}
    }

    \begin{subfigure}[t]{0.48\linewidth}
        \centering
        \resizebox{0.95\linewidth}{!}{
        \begin{tikzpicture}
            \tikzset{
                env/.style={rectangle, draw=black!80, line width=1.5pt, fill=black!10, minimum size=1cm, font=\huge\bfseries},
                var/.style={circle, draw=black!90, line width=1.5pt, fill=white, font=\huge\bfseries, minimum size=1.2cm},
                latent/.style={circle, draw=gray!80, dashed, line width=1.5pt, fill=gray!10, font=\huge\bfseries, minimum size=1.2cm},
                var_latent/.style={circle, draw=blue!70, dashed, line width=1.5pt, fill=blue!5, font=\huge\bfseries, minimum size=1.2cm}
            }
            \useasboundingbox (-5.5, 3.5) rectangle (3.5, -4.0);
            \node[var_latent] (Xz) at (0, 1.8) {$X^{\perp}_Z$};
            \node[var_latent] (Xy) at (0, -1.8) {$X^{\perp}_Y$};
            \node[latent] (Z) at (-2.5, 0) {$Z$};
            \node[var] (Y) at (2.5, 0) {$Y$};
            \node[env] (E) at (-4.2, 0) {$E$};
            \draw[arrow, line width=1.5pt] (Xz) -- (Y);
            \draw[arrow, line width=1.5pt] (Z) -- (Xy);
            \draw[arrow, bend left=90, looseness=1.5, line width=1.5pt] (Y) to (Z);
            \draw[arrow, line width=1.5pt] (E) -- (Z);
            \begin{scope}[on background layer]
                \node[observed_box, fit=(Xz) (Xy), inner sep=10pt] (TextBox) {};
            \end{scope}
            \node[text=red!80, font=\large\bfseries, anchor=south, yshift=4pt] at (TextBox.north) {Observed Text ($X$)};
        \end{tikzpicture}
        }
        \caption{Causal setting}
    \end{subfigure}
    \hfill
    \begin{subfigure}[t]{0.48\linewidth}
        \centering
        \resizebox{0.95\linewidth}{!}{
        \begin{tikzpicture}
            \tikzset{
                env/.style={rectangle, draw=black!80, line width=1.5pt, fill=black!10, minimum size=1cm, font=\huge\bfseries},
                var/.style={circle, draw=black!90, line width=1.5pt, fill=white, font=\huge\bfseries, minimum size=1.2cm},
                latent/.style={circle, draw=gray!80, dashed, line width=1.5pt, fill=gray!10, font=\huge\bfseries, minimum size=1.2cm},
                var_latent/.style={circle, draw=blue!70, dashed, line width=1.5pt, fill=blue!5, font=\huge\bfseries, minimum size=1.2cm}
            }
            \useasboundingbox (-5.5, 3.5) rectangle (3.5, -4.0);
            \node[var_latent] (Xz) at (0, 1.8) {$X^{\perp}_Z$};
            \node[var_latent] (Xy) at (0, -1.8) {$X^{\perp}_Y$};
            \node[latent] (Z) at (-2.5, 0) {$Z$};
            \node[var] (Y) at (2.5, 0) {$Y$};
            \node[env] (E) at (-4.2, 0) {$E$};
            \draw[arrow, line width=1.5pt] (Y) -- (Xz);
            \draw[arrow, line width=1.5pt] (Z) -- (Xy);
            \draw[arrow, bend left=90, looseness=1.5, line width=1.5pt] (Y) to (Z);
            \draw[arrow, line width=1.5pt] (E) -- (Z);
            \begin{scope}[on background layer]
                \node[observed_box, fit=(Xz) (Xy), inner sep=10pt] (TextBox) {};
            \end{scope}
            \node[text=red!80, font=\large\bfseries, anchor=south, yshift=4pt] at (TextBox.north) {Observed Text ($X$)};
        \end{tikzpicture}
        }
        \caption{Anti-causal setting}
    \end{subfigure}
    \caption{\textbf{Label-Generated Feature.} Spurious mechanism: $Y \rightarrow Z \rightarrow X^\perp_Y$. The red box indicates the observable text $X$.}
    \label{fig:label-generated}
\end{figure}
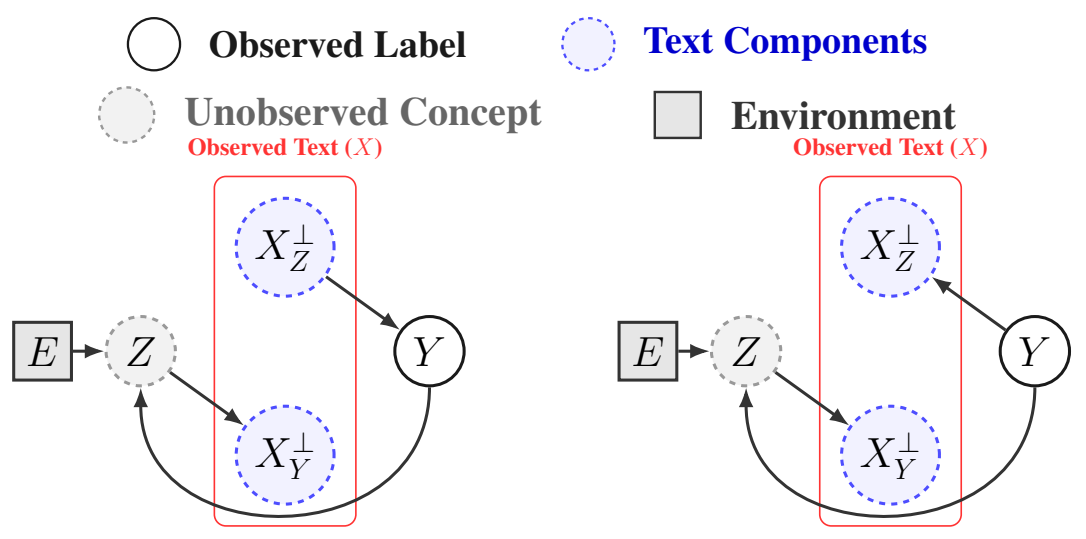

\begin{figure}[!htbp]
    \centering
    \begin{subfigure}[t]{0.48\linewidth}
        \centering
        \resizebox{0.95\linewidth}{!}{
        \begin{tikzpicture}
            \tikzset{
                env/.style={rectangle, draw=black!80, line width=1.5pt, fill=black!10, minimum size=1cm, font=\huge\bfseries},
                var/.style={circle, draw=black!90, line width=1.5pt, fill=white, font=\huge\bfseries, minimum size=1.2cm},
                latent/.style={circle, draw=gray!80, dashed, line width=1.5pt, fill=gray!10, font=\huge\bfseries, minimum size=1.2cm},
                var_latent/.style={circle, draw=blue!70, dashed, line width=1.5pt, fill=blue!5, font=\huge\bfseries, minimum size=1.2cm},
                arrow/.style={-latex, line width=1.5pt}
            }
            \useasboundingbox (-5.5, 3.5) rectangle (3.5, -4.0);
            \node[var_latent] (Xz) at (0, 1.8) {$X^{\perp}_Z$};
            \node[var_latent] (Xy) at (0, -1.8) {$X^{\perp}_Y$};
            \node[latent] (Z) at (-2.5, 0) {$Z$};
            \node[latent] (U) at (0, -3.2) {$U$};
            \node[var] (Y) at (2.5, 0) {$Y$};
            \node[env] (E) at (-4.2, 0) {$E$};
            \draw[arrow] (Z) -- (Xy);
            \draw[arrow] (Xz) -- (Y);
            \draw[arrow] (E) -- (Z);
            \draw[arrow, bend left=35] (U) to (Z);
            \draw[arrow, bend right=30] (U) to (Y);
            \begin{scope}[on background layer]
                \node[observed_box, fit=(Xz) (Xy), inner sep=10pt] (TextBox) {};
            \end{scope}
            \node[text=red!80, font=\large\bfseries, anchor=south, yshift=4pt] at (TextBox.north) {Observed Text ($X$)};
        \end{tikzpicture}
        }
        \caption{Causal setting}
        \label{fig:causal_confounding}
    \end{subfigure}
    \hfill
    \begin{subfigure}[t]{0.48\linewidth}
        \centering
        \resizebox{0.95\linewidth}{!}{
        \begin{tikzpicture}
            \tikzset{
                env/.style={rectangle, draw=black!80, line width=1.5pt, fill=black!10, minimum size=1cm, font=\huge\bfseries},
                var/.style={circle, draw=black!90, line width=1.5pt, fill=white, font=\huge\bfseries, minimum size=1.2cm},
                latent/.style={circle, draw=gray!80, dashed, line width=1.5pt, fill=gray!10, font=\huge\bfseries, minimum size=1.2cm},
                var_latent/.style={circle, draw=blue!70, dashed, line width=1.5pt, fill=blue!5, font=\huge\bfseries, minimum size=1.2cm},
                arrow/.style={-latex, line width=1.5pt}
            }
            \useasboundingbox (-5.5, 3.5) rectangle (3.5, -4.0);
            \node[var_latent] (Xz) at (0, 1.8) {$X^{\perp}_Z$};
            \node[var_latent] (Xy) at (0, -1.8) {$X^{\perp}_Y$};
            \node[latent] (Z) at (-2.5, 0) {$Z$};
            \node[latent] (U) at (0, -3.2) {$U$};
            \node[var] (Y) at (2.5, 0) {$Y$};
            \node[env] (E) at (-4.2, 0) {$E$};
            \draw[arrow] (Z) -- (Xy);
            \draw[arrow] (Y) -- (Xz); 
            \draw[arrow] (E) -- (Z);
            \draw[arrow, bend left=35] (U) to (Z);
            \draw[arrow, bend right=30] (U) to (Y);
            \begin{scope}[on background layer]
                \node[observed_box, fit=(Xz) (Xy), inner sep=10pt] (TextBox) {};
            \end{scope}
            \node[text=red!80, font=\large\bfseries, anchor=south, yshift=4pt] at (TextBox.north) {Observed Text ($X$)};
        \end{tikzpicture}
        }
        \caption{Anti-causal setting}
        \label{fig:anti_confounding}
    \end{subfigure}
    \caption{\textbf{Confounding.} Backdoor path via unobserved $U$.}
    \label{fig:confounding}
\end{figure}
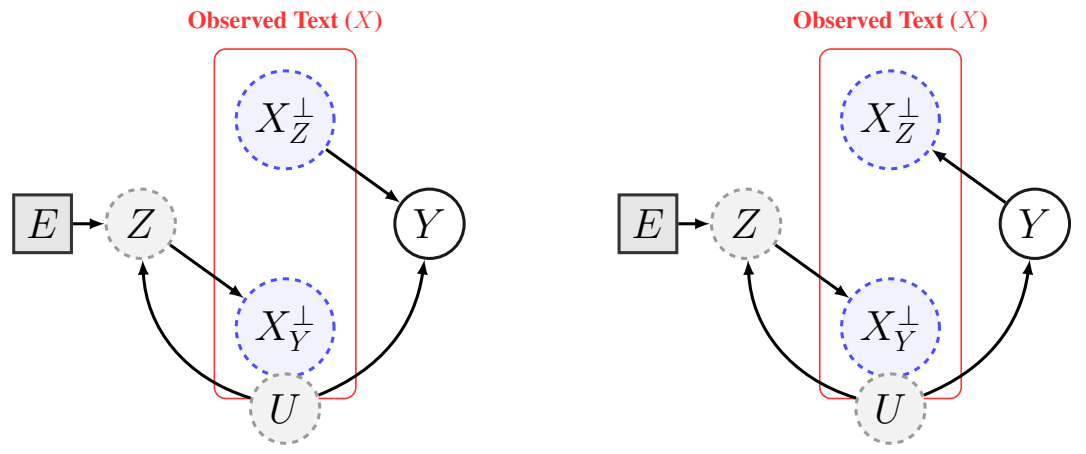

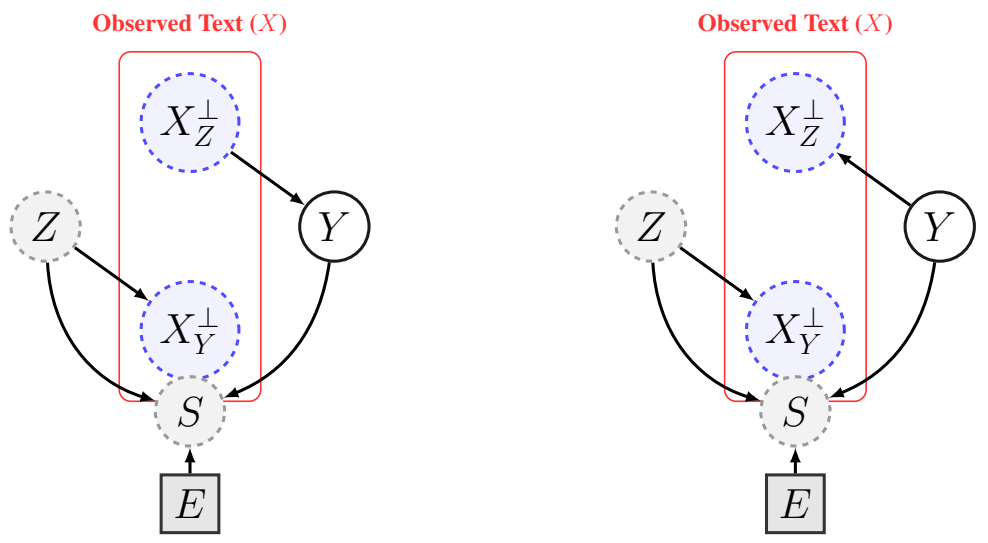
\begin{figure}[!htbp]
    \centering
    \begin{subfigure}[t]{0.48\linewidth}
        \centering
        \resizebox{0.95\linewidth}{!}{
        \begin{tikzpicture}
            \tikzset{
                env/.style={rectangle, draw=black!80, line width=1.5pt, fill=black!10, minimum size=1cm, font=\huge\bfseries},
                var/.style={circle, draw=black!90, line width=1.5pt, fill=white, font=\huge\bfseries, minimum size=1.2cm},
                latent/.style={circle, draw=gray!80, dashed, line width=1.5pt, fill=gray!10, font=\huge\bfseries, minimum size=1.2cm},
                selection/.style={circle, draw=gray!80, dashed, line width=1.5pt, fill=gray!10, minimum size=1.2cm, font=\huge\bfseries},
                var_latent/.style={circle, draw=blue!70, dashed, line width=1.5pt, fill=blue!5, font=\huge\bfseries, minimum size=1.2cm},
                arrow/.style={-latex, line width=1.5pt}
            }
            \useasboundingbox (-5.5, 3.5) rectangle (3.5, -5.5);
            \node[var_latent] (Xz) at (0, 1.8) {$X^{\perp}_Z$};
            \node[var_latent] (Xy) at (0, -1.8) {$X^{\perp}_Y$};
            \node[latent] (Z) at (-2.5, 0) {$Z$};
            \node[selection] (S) at (0, -3.2) {$S$};
            \node[var] (Y) at (2.5, 0) {$Y$};
            \node[env] (E) at (0, -4.8) {$E$};
            \draw[arrow] (Z) -- (Xy);
            \draw[arrow] (Xz) -- (Y);
            \draw[arrow, bend right=35] (Z) to (S);
            \draw[arrow, bend left=30]  (Y) to (S);
            \draw[arrow] (E) -- (S);
            \begin{scope}[on background layer]
                \node[observed_box, fit=(Xz) (Xy), inner sep=10pt] (TextBox) {};
            \end{scope}
            \node[text=red!80, font=\large\bfseries, anchor=south, yshift=4pt] at (TextBox.north) {Observed Text ($X$)};
        \end{tikzpicture}
        }
        \caption{Causal setting}
        \label{fig:causal_selection}
    \end{subfigure}
    \hfill
    \begin{subfigure}[t]{0.48\linewidth}
        \centering
        \resizebox{0.95\linewidth}{!}{
        \begin{tikzpicture}
            \tikzset{
                env/.style={rectangle, draw=black!80, line width=1.5pt, fill=black!10, minimum size=1cm, font=\huge\bfseries},
                var/.style={circle, draw=black!90, line width=1.5pt, fill=white, font=\huge\bfseries, minimum size=1.2cm},
                latent/.style={circle, draw=gray!80, dashed, line width=1.5pt, fill=gray!10, font=\huge\bfseries, minimum size=1.2cm},
                selection/.style={circle, draw=gray!80, dashed, line width=1.5pt, fill=gray!10, minimum size=1.2cm, font=\huge\bfseries},
                var_latent/.style={circle, draw=blue!70, dashed, line width=1.5pt, fill=blue!5, font=\huge\bfseries, minimum size=1.2cm},
                arrow/.style={-latex, line width=1.5pt}
            }
            \useasboundingbox (-5.5, 3.5) rectangle (3.5, -5.5);
            \node[var_latent] (Xz) at (0, 1.8) {$X^{\perp}_Z$};
            \node[var_latent] (Xy) at (0, -1.8) {$X^{\perp}_Y$};
            \node[latent] (Z) at (-2.5, 0) {$Z$};
            \node[selection] (S) at (0, -3.2) {$S$};
            \node[var] (Y) at (2.5, 0) {$Y$};
            \node[env] (E) at (0, -4.8) {$E$};
            \draw[arrow] (Z) -- (Xy);
            \draw[arrow] (Y) -- (Xz); 
            \draw[arrow, bend right=35] (Z) to (S);
            \draw[arrow, bend left=30]  (Y) to (S);
            \draw[arrow] (E) -- (S);
            \begin{scope}[on background layer]
                \node[observed_box, fit=(Xz) (Xy), inner sep=10pt] (TextBox) {};
            \end{scope}
            \node[text=red!80, font=\large\bfseries, anchor=south, yshift=4pt] at (TextBox.north) {Observed Text ($X$)};
        \end{tikzpicture}
        }
        \caption{Anti-causal setting}
        \label{fig:anti_selection}
    \end{subfigure}
    \caption{\textbf{Selection Bias.} Conditioned on selection collider $S$.}
    \label{fig:selection-bias}
\end{figure}

Controlled shortcut injection makes it possible to manipulate a feature's correlation with the label and directly measure the consequences of relying on it. Previous work has used this approach to study shortcut formation and evaluate model explanations \citep{bastings2022will, pezeshkpour2022combining}. Following the environment-based formulation of IRM \citep{arjovsky2019invariant}, we construct two training environments in which a designated shortcut remains predictive but varies in strength, together with an OOD test environment in which its correlation with the label is inverted. Unlike prior protocols that treat injected artifacts as a single class of spurious correlations, we organize our environments according to the process generating the shortcut--label correlation. Following the causal terminology of \citet{veitch2021counterfactual}, we distinguish label-generated features, confounding, and selection bias, using graphical representations inspired by their framework. For each mechanism, we vary only the structural equation or sampling component responsible for the shortcut while preserving the task-relevant relationship. Table~\ref{tab:params} summarizes the shared environment parameters, while Figures~\ref{fig:label-generated}, \ref{fig:confounding}, and \ref{fig:selection-bias} illustrate the corresponding structures.

\begin{table}[t]
\centering
\small
\renewcommand{\arraystretch}{1.15}
\setlength{\tabcolsep}{4pt}
\begin{tabular}{@{}lcccc@{}}
\toprule
\textbf{Mechanism} & \textbf{Param.} & \textbf{$e_1$} & \textbf{$e_2$} & \textbf{OOD} \\
\midrule
Label-generated & $p_c$ & $0.70$ & $0.90$ & $0.00$ \\
Confounding & $a$ & $0.01$ & $0.10$ & $1.00$ \\
Selection bias & $p_s$ & $0.70$ & $0.90$ & $0.00$ \\
\bottomrule
\end{tabular}
\caption{Environment parameters shared across datasets. The OOD environment inverts the training-time shortcut. Parameter definitions and dataset-specific shortcut realizations are provided in Appendix~\ref{app:mechanisms}.}
\label{tab:params}
\end{table}

\paragraph{Label-Generated Feature (Figure~\ref{fig:label-generated}).} In this mechanism, $Y \rightarrow Z \rightarrow X_Y^{\perp}$: the target label $Y$ generates a latent concept $Z$, which in turn produces the observable spurious feature $X_Y^{\perp}$. Mirroring the classic Colored MNIST setup \citet{arjovsky2019invariant}, this label-dependent process can make the shortcut more predictive than the genuine task-relevant textual signal.

We implement this mechanism by inserting a class-associated token before neutral function words\footnote{For example, \emph{the, a, an, of, in, to}.} throughout the text. The parameter $p_c^{(e)} = P^{(e)}\!\left(Z=Y\mid Y\right)$ controls the probability that the injected token agrees with the label. The OOD intervention reverses this correlation. To prevent the task-relevant signal from trivially dominating, we apply label noise at rate $\epsilon=0.2$ before token injection.

\paragraph{Confounding (Figure~\ref{fig:confounding}).} An unobserved confounder $U$ influences both the target $Y$ and a latent concept $Z$, opening the backdoor path $Y \leftarrow U \rightarrow Z \rightarrow X_Y^{\perp}$. This induces a non-causal statistical dependency that models can exploit as a predictive shortcut. The influence of $U$ on $Y$ remains fixed across environments, while only the fidelity of the proxy $Z$ varies. This preserves the invariance conditions required by IRM, see Appendices~\ref{app:sem} and~\ref{app:confounding-validity}. 

Experimentally, we sample a binary confounder $U\sim\operatorname{Ber}(0.5)$, which influences $Y$ through a fixed, environment-independent mechanism. Its environment-dependent proxy is generated as $Z^{(e)} = U \oplus \operatorname{Ber}\!\left(a^{(e)}\right)$, where $\oplus$ denotes XOR. The parameter $a^{(e)}$ is the probability of flipping the value of $U$, so smaller values produce a more faithful shortcut. The OOD intervention completely reverses the proxy while leaving the relationship between $U$ and $Y$ unchanged. The resulting concept $Z^{(e)}$ determines the injected textual feature.

\paragraph{Selection Bias (Figure~\ref{fig:selection-bias}).} An unobserved selection mechanism $S$ is influenced by both $Y$ and a latent feature $Z$, forming the collider $Y \rightarrow S \leftarrow Z$. The feature $Z$ also generates the observable component $X_Y^{\perp}$. Training only on examples satisfying $S=1$ induces a non-causal correlation between $Y$ and $X_Y^{\perp}$. Unlike the previous mechanisms, selection does not change their underlying values, but biases their observed co-occurrence frequencies.

We simulate selection bias by subsampling each dataset according to the alignment between $Y$ and a dataset-specific latent feature $Z$. Examples for which $Z$ heuristically agrees with $Y$ are called \emph{aligned}, whereas the remaining examples are \emph{misaligned}. The parameter $p_s^{(e)} = P^{(e)}\!\left(\text{aligned}\mid S=1\right)$ controls the proportion of aligned examples retained in environment $e$. The OOD test set contains only misaligned examples, making the training-time correlation systematically misleading. The dataset-specific definitions of $Z$ and alignment are provided in Appendix~\ref{app:mechanisms}.

\subsection{Datasets}
\label{sec:datasets}

We evaluate the three shortcut-generating mechanisms on three balanced text-classification benchmarks covering causal and anti-causal tasks. Each corpus is partitioned into 80\% training, 10\% in-distribution validation, and 10\% OOD test data. The training partition is divided equally between the two environments defined in Section~\ref{sec:environment-construction}.

\paragraph{Amazon Books Reviews \citep{hou2026bridging} (Causal, $X \rightarrow Y$):} We predict whether a book review is unhelpful ($Y=0$) or helpful ($Y=1$) based on its text and user votes. The balanced corpus contains 100,000 reviews, including 80,000 training examples. We treat this task as causal because the review's informativeness and quality influence its perceived helpfulness.

\paragraph{AG News \citep{zhang2015characterlevel} (Causal, $X \rightarrow Y$):} We classify news articles into World, Sports, Business, or Sci/Tech ($Y\in\{0,1,2,3\}$). The balanced corpus contains 127,600 articles, including 101,460 training examples. We treat this task as causal because the linguistic content of an article determines its thematic category.

\paragraph{IMDB Genres \citep{maas-etal-2011-learning} (Anti-causal, $Y \rightarrow X$):} We classify movie plot descriptions as Thriller ($Y=0$) or Romance ($Y=1$). The balanced corpus contains 87,533 descriptions, including 70,027 training examples. We treat the task as anti-causal because the movie genre acts as a latent generative variable that influences the description's vocabulary, style, and thematic content. Although IRM is often motivated through causal prediction, it can also identify invariant predictive relationships in anti-causal settings when class proportions remain stable across environments, as ensured by our balanced construction. Appendix~\ref{app:causal-direction} provides the formal argument and discusses the limitation under label shift.

\section{Results}

We evaluate the ability of IRM to identify robust predictors across nine experimental configurations (three spurious correlation mechanisms applied to three datasets, as summarized in Table~\ref{tab:params}). The aim is to quantify the robustness of invariance against the structural nature of the misleading correlation and environmental variability.

\subsection{ID and OOD Performance}

Tables~\ref{tab:id-accuracy} and~\ref{tab:ood-accuracy} report, respectively, the in-distribution (ID) validation accuracy and the OOD test accuracy across all configurations.

\begin{table*}[!htbp]
\centering
\small
\begin{tabular}{@{}lccc@{}}
\toprule
\textbf{Spurious Mechanism} & \textbf{Amazon} (Causal) & \textbf{AG News} (Causal) & \textbf{IMDB} (Anti-causal) \\ \midrule
\multirow{2}{*}{\textbf{Label-Generated}}
  & \ms{0.814}{0.001} / \ms{0.758}{0.008}
  & \ms{0.810}{0.000} / \ms{0.769}{0.004}
  & \ms{0.812}{0.002} / \ms{0.790}{0.007} \\
  & (\textcolor{red}{$-0.056$}) & (\textcolor{red}{$-0.041$}) & (\textcolor{red}{$-0.022$}) \\ \midrule
\multirow{2}{*}{\textbf{Confounding}}
  & \ms{0.869}{0.005} / \ms{0.833}{0.014}
  & \ms{0.862}{0.003} / \ms{0.745}{0.002}
  & \ms{0.740}{0.002} / \ms{0.718}{0.002} \\
  & (\textcolor{red}{$-0.036$}) & (\textcolor{red}{$-0.117$}) & (\textcolor{red}{$-0.022$}) \\ \midrule
\multirow{2}{*}{\textbf{Selection Bias}}
  & \ms{0.843}{0.003} / \ms{0.788}{0.009}
  & \ms{0.926}{0.001} / \ms{0.881}{0.004}
  & \ms{0.794}{0.003} / \ms{0.716}{0.005} \\
  & (\textcolor{red}{$-0.055$}) & (\textcolor{red}{$-0.045$}) & (\textcolor{red}{$-0.078$}) \\ \bottomrule
\end{tabular}
\caption{In-distribution (ID) validation accuracy for ERM and IRM.}
\label{tab:id-accuracy}
\end{table*}

As expected in Table~\ref{tab:id-accuracy}, ERM achieves higher ID accuracy in all configurations, with IRM incurring losses of $2.2$ to $11.7$ percentage points. This reflects the cost of discarding unstable features that remain predictive within the training distribution.

\begin{table*}[!htbp]
\centering
\small
\begin{tabular}{@{}lccc@{}}
\toprule
\textbf{Spurious Mechanism} & \textbf{Amazon} (Causal) & \textbf{AG News} (Causal) & \textbf{IMDB} (Anti-causal) \\ \midrule
\multirow{2}{*}{\textbf{Label-Generated}}
  & \ms{0.117}{0.003} / \ms{0.672}{0.007}
  & \ms{0.417}{0.011} / \ms{0.798}{0.007}
  & \ms{0.147}{0.009} / \ms{0.728}{0.008} \\
  & (\textcolor{blue}{$+0.555$}) & (\textcolor{blue}{$+0.381$}) & (\textcolor{blue}{$+0.581$}) \\ \midrule
\multirow{2}{*}{\textbf{Confounding}}
  & \ms{0.501}{0.006} / \ms{0.638}{0.032}
  & \ms{0.244}{0.010} / \ms{0.699}{0.014}
  & \ms{0.622}{0.005} / \ms{0.691}{0.012} \\
  & (\textcolor{blue}{$+0.137$}) & (\textcolor{blue}{$+0.455$}) & (\textcolor{blue}{$+0.069$}) \\ \midrule
\multirow{2}{*}{\textbf{Selection Bias}}
  & \ms{0.544}{0.005} / \ms{0.721}{0.004}
  & \ms{0.756}{0.008} / \ms{0.836}{0.010}
  & \ms{0.209}{0.006} / \ms{0.701}{0.025} \\
  & (\textcolor{blue}{$+0.177$}) & (\textcolor{blue}{$+0.080$}) & (\textcolor{blue}{$+0.492$}) \\ \bottomrule
\end{tabular}
\caption{Out-of-distribution (OOD) test accuracy for ERM and IRM.}
\label{tab:ood-accuracy}
\end{table*}

Under shortcut inversion, IRM outperforms ERM in all nine dataset--mechanism configurations, with gains ranging from $6.9$ to $58.1$ percentage points. The improvement spans label-generated features, confounding, and selection bias, suggesting that shortcut rejection does not depend on a single generating mechanism. It also holds on the anti-causal IMDB task under the stable class proportions imposed by our balanced construction, consistent with the conditions discussed in Appendix~\ref{app:causal-direction}.

\subsection{Sensitivity Analysis}
\label{sec:sensitivity-analysis}

To identify the minimal conditions for IRM to succeed, we varied the statistical disparity between training environments, defined as the gap in the strength of the spurious correlation across $\mathcal{E}_{\text{tr}}$ \citep{choe2020empirical}. This disparity is the fundamental requirement for the invariance penalty: it creates the distributional instability that allows the model to distinguish unreliable shortcuts from stable causal correlation. Concretely, for a target gap $\Delta$, environments are constructed symmetrically around a fixed center $c$: $p_1 = c - \Delta/2$ and $p_2 = c + \Delta/2$. For Label-Generated and Selection Bias, we fix $c=0.85$ so that both environments maintain a globally strong spurious signal; only their differential varies (e.g., $\Delta=0.20$ yields $p_1=0.75$, $p_2=0.95$). For Confounding, the center is $c=0.05$ on the proxy-noise parameter $a$, with $\Delta \in [0, 0.18]$. At $\Delta=0$, the two environments are statistically identical and IRM cannot detect any correlation instability, degenerating to ERM.

\begin{figure}[t]
\centering
\begin{subfigure}[t]{0.48\linewidth}
\includegraphics[width=\linewidth]{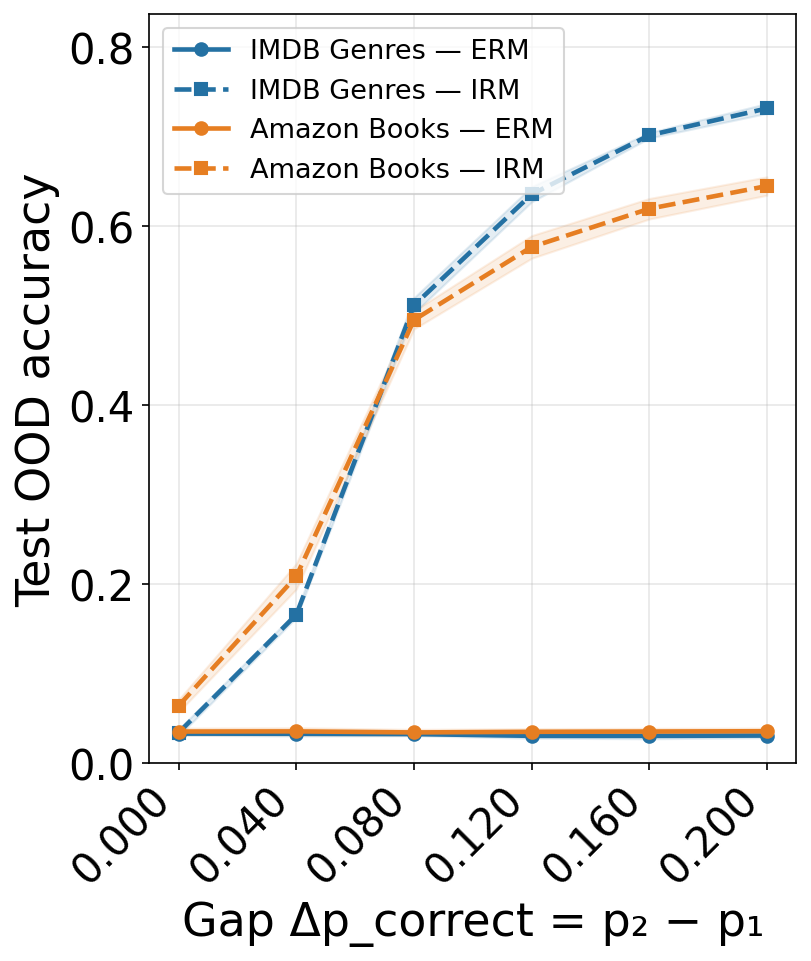}
\caption{Label-Generated}
\end{subfigure}
\hfill
\begin{subfigure}[t]{0.48\linewidth}
\includegraphics[width=\linewidth]{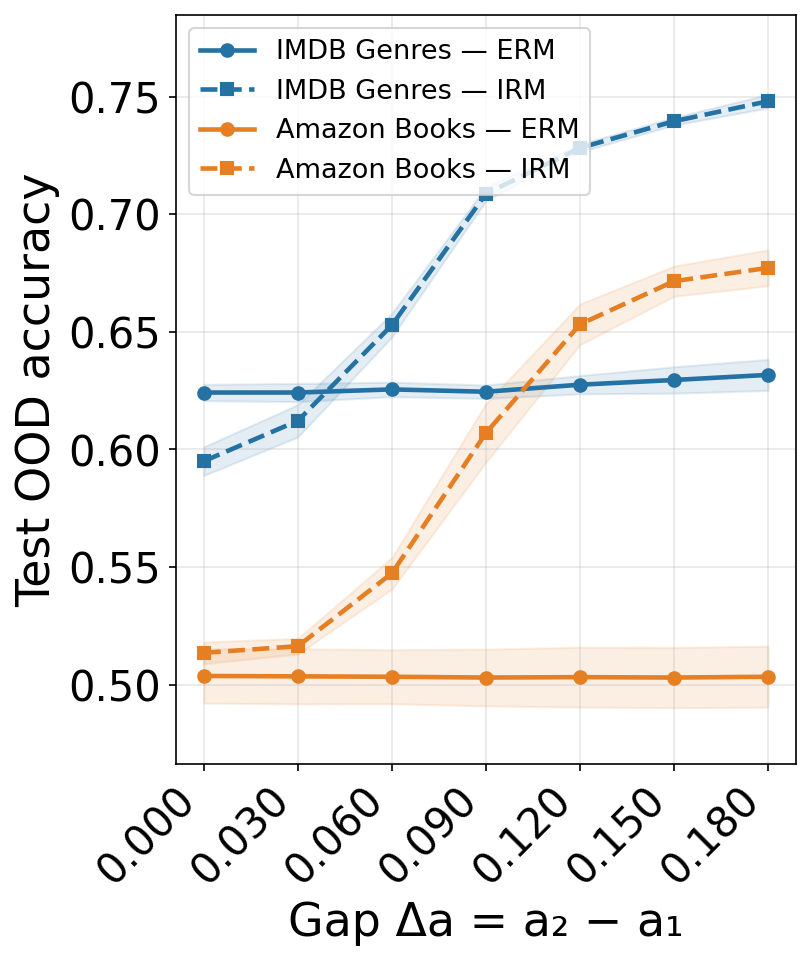}
\caption{Confounding}
\end{subfigure}
\vskip 4pt
\begin{subfigure}[t]{0.48\linewidth}
\includegraphics[width=\linewidth]{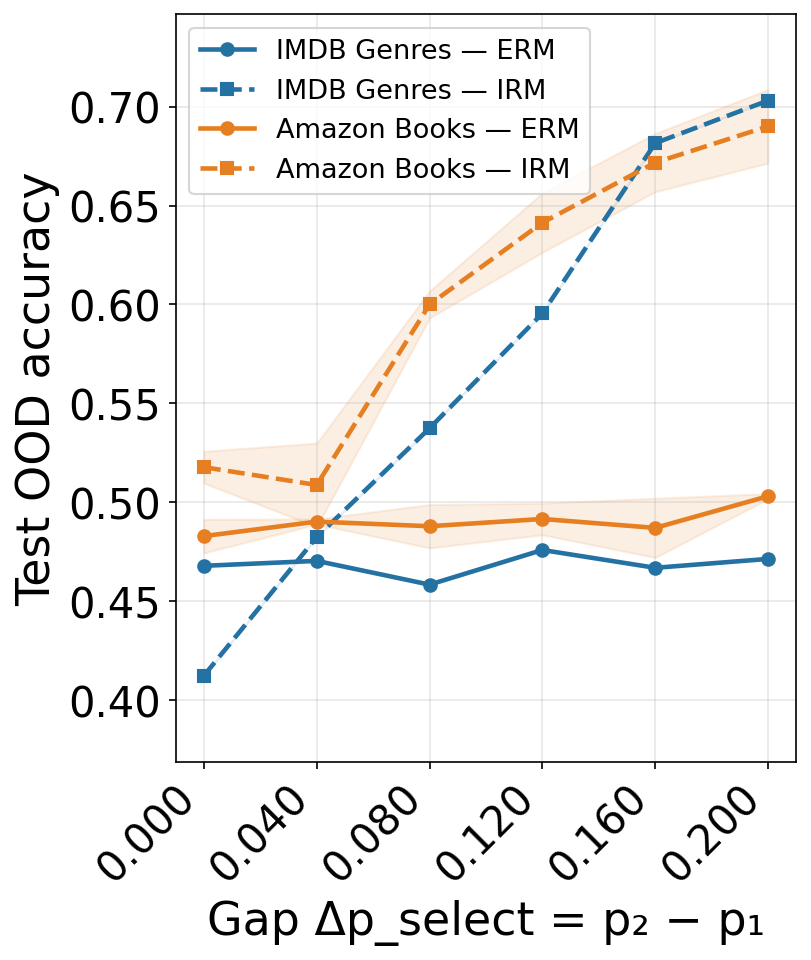}
\caption{Selection Bias}
\end{subfigure}
\caption{\textbf{Sensitivity analysis.} OOD accuracy as a function of the correlation gap for label-generated features (top-left), confounding (top-right), and selection bias (bottom-center).}
\label{fig:sensitivity}
\end{figure}

Our results (Figure~\ref{fig:sensitivity}) reveal that IRM can leverage even the slightest statistical disparity. While a minimal gap is often sufficient for the invariance penalty to begin discarding the misleading correlation in favor of a stable task-relevant signal, we observe that IRM's OOD performance tends to increase as the correlation gap between environments widens. This sensitivity is an encouraging sign for real-world applications where domain differences may be subtle.

\subsection{Robustness to Label Noise}
\label{sec:noise-sweep}

We investigate the models' behavior as the reliability of the task-relevant signal is progressively degraded. By gradually increasing the label noise rate $\epsilon$, we artificially weaken the predictive power of the causal features relative to the spurious ones. This experiment is designed to determine whether a model continues to favor invariant features as the task-relevant signal becomes less reliable.

\begin{figure*}[t]
\centering
\begin{subfigure}[t]{0.48\linewidth}
\includegraphics[width=\linewidth]{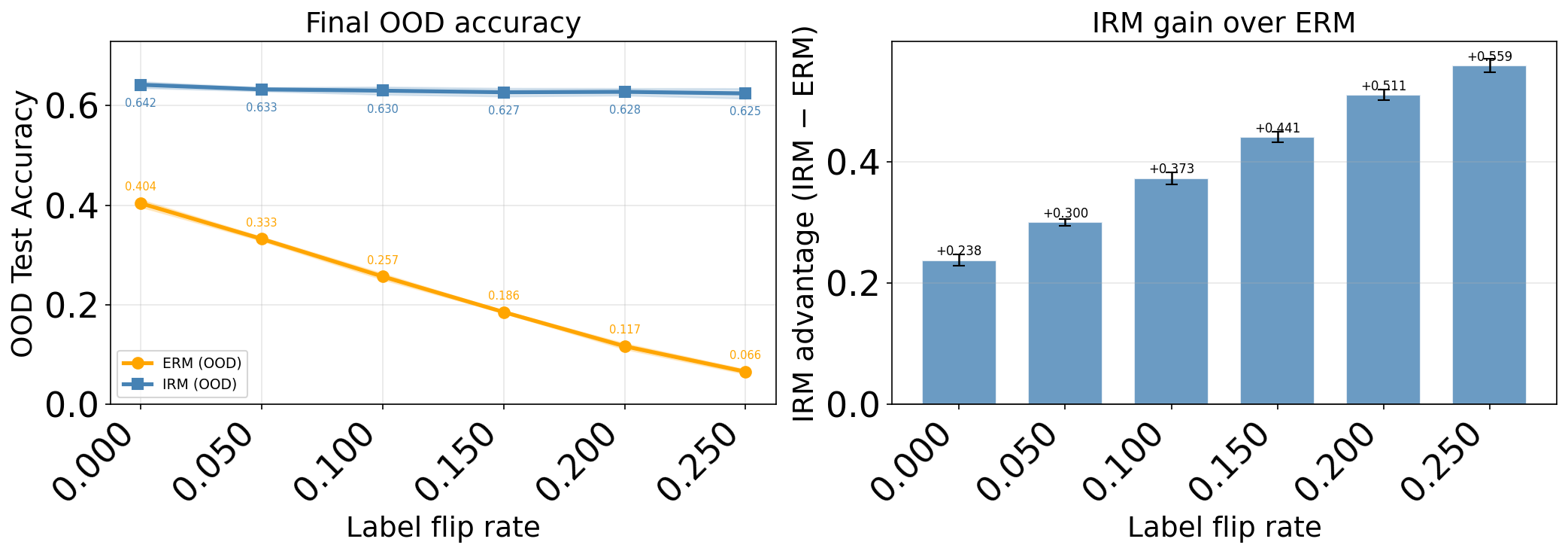}
\caption{Amazon Books --- Label-Generated}
\end{subfigure}
\hfill
\begin{subfigure}[t]{0.48\linewidth}
\includegraphics[width=\linewidth]{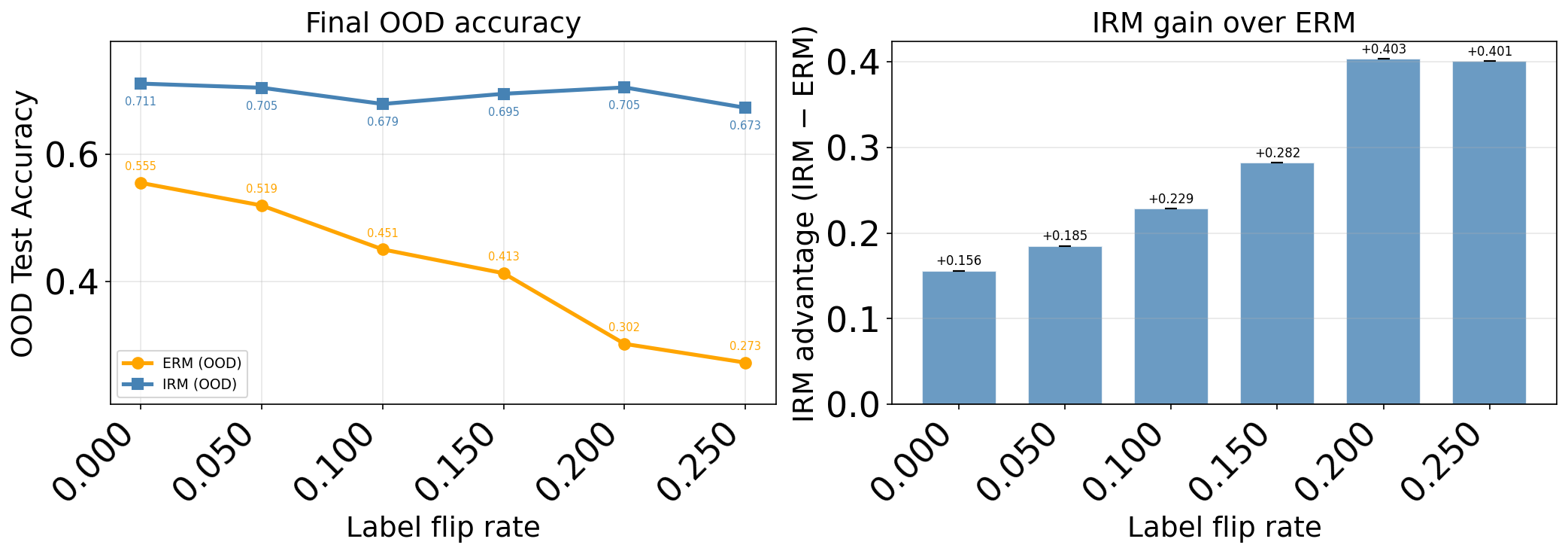}
\caption{Amazon Books --- Selection Bias}
\end{subfigure}
\vskip 4pt
\begin{subfigure}[t]{0.48\linewidth}
\includegraphics[width=\linewidth]{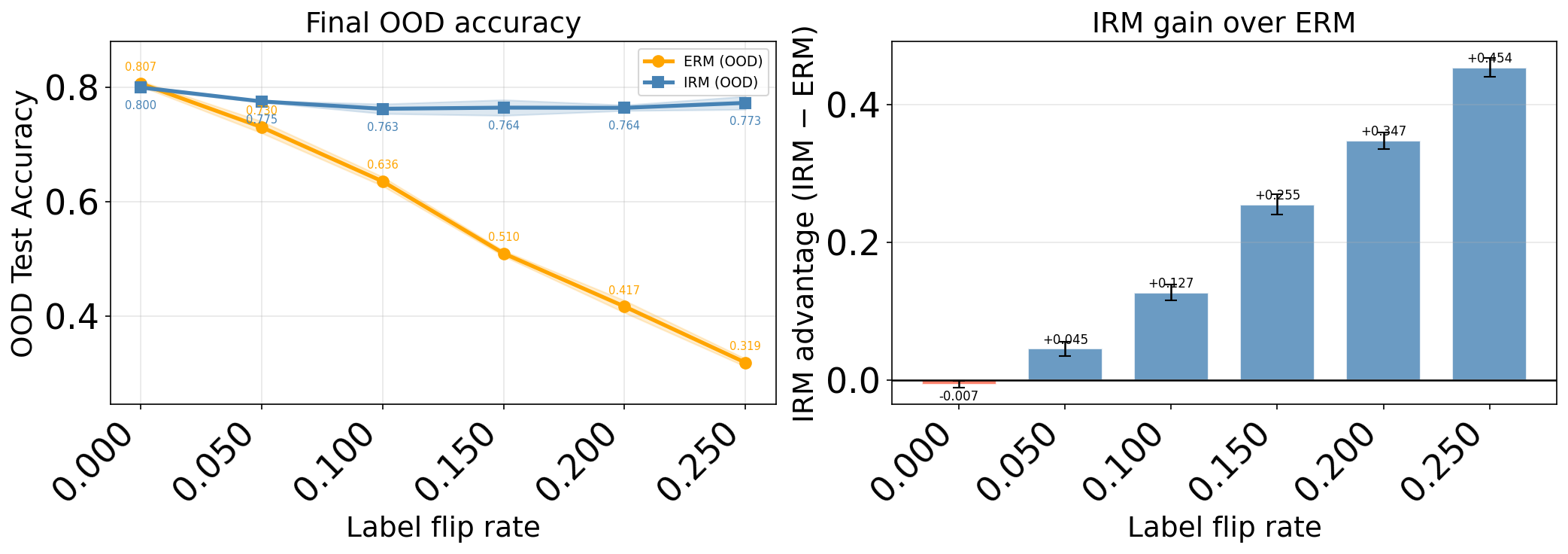}
\caption{AG News --- Label-Generated}
\end{subfigure}
\hfill
\begin{subfigure}[t]{0.48\linewidth}
\includegraphics[width=\linewidth]{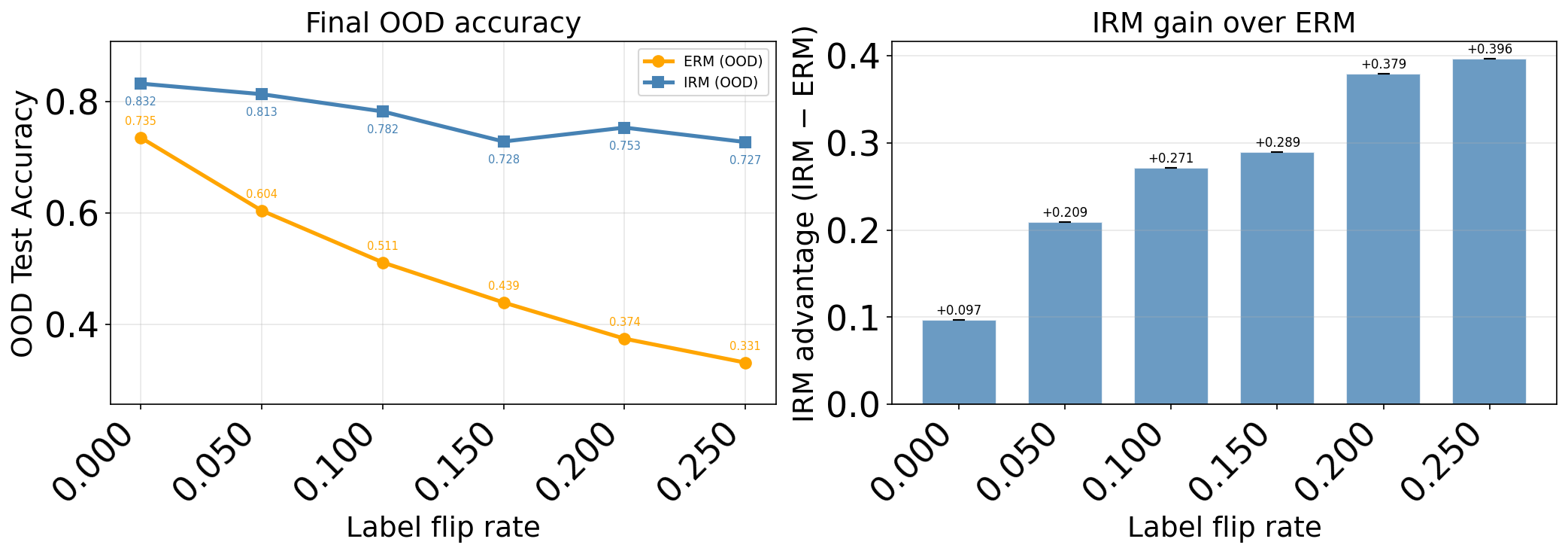}
\caption{AG News --- Selection Bias}
\end{subfigure}
\vskip 4pt
\begin{subfigure}[t]{0.48\linewidth}
\includegraphics[width=\linewidth]{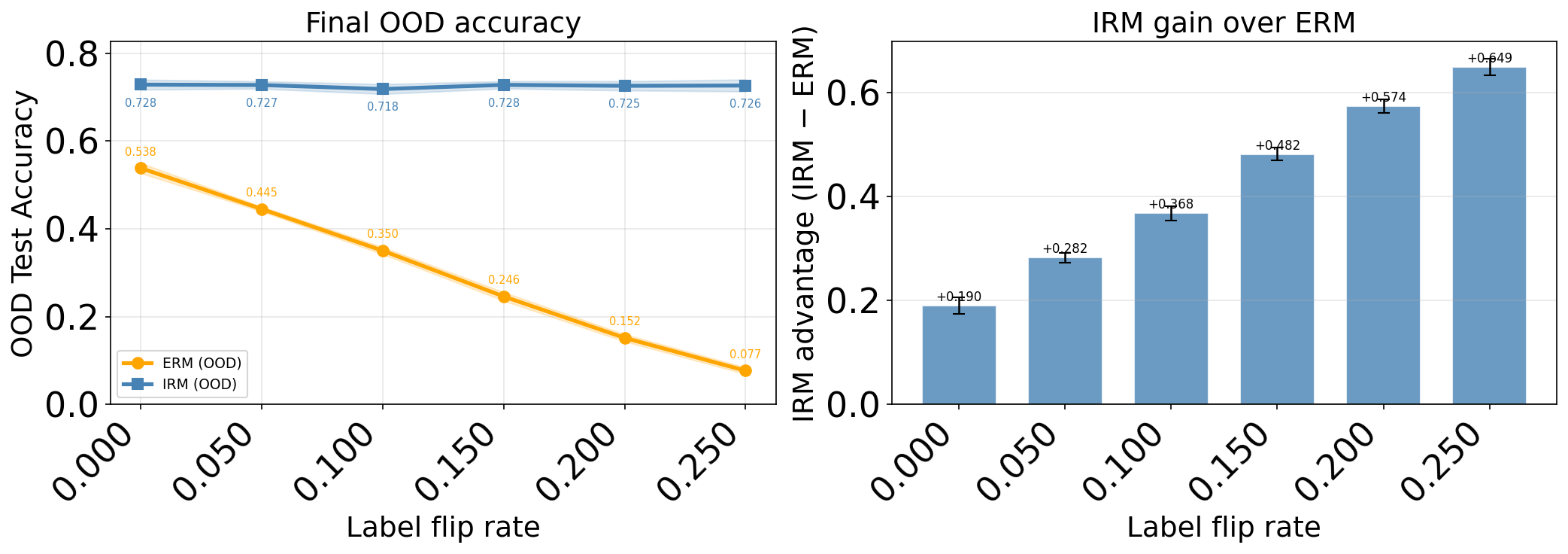}
\caption{IMDB Genres --- Label-Generated}
\end{subfigure}
\hfill
\begin{subfigure}[t]{0.48\linewidth}
\includegraphics[width=\linewidth]{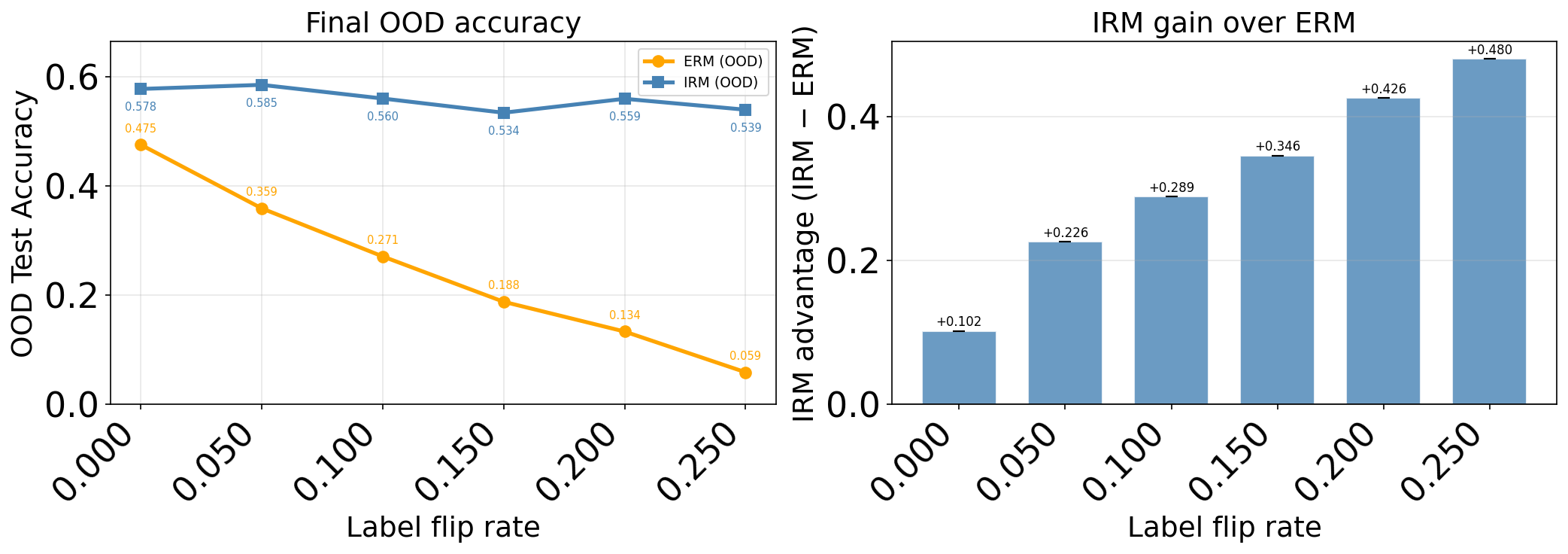}
\caption{IMDB Genres --- Selection Bias}
\end{subfigure}
\caption{\textbf{Noise-sweep analysis.} OOD accuracy as a function of label noise. Rows correspond to Amazon Books (top), AG News (middle), and IMDB Genres (bottom). Columns correspond to label-generated features (left) and selection bias (right).}
\label{fig:noise-sweep}
\end{figure*}

As shown in Figure~\ref{fig:noise-sweep}, increasing $\epsilon$ progressively weakens the causal signal while leaving the shortcut configuration unchanged, making the spurious signal relatively more attractive for minimizing training loss. ERM succumbs to this bias. As the genuine signal fades, the model shifts its reliance to the shortcut, leading to catastrophic performance degradation at test time when the correlation is inverted. Conversely, IRM remains largely insensitive to label noise. IRM does not select features based on their relative predictive convenience, but strictly based on their structural stability. By rejecting correlations that vary across environments, IRM maintains its focus on the causal features, preferring a noisy yet invariant signal over a cleaner but unstable shortcut, even when the latter offers lower training loss.

\section{Beyond Synthetic Shortcuts}
\label{sec:beyond-synthetic}

The controlled experiments in Section~\ref{sec:experiment-design} isolate the origin and strength of each spurious correlation. We now consider two less controlled settings. Bias in Bios \citep{de-arteaga2019bias} contains a naturally occurring association between gender and profession, whereas ActorCorr \citep{eshuijs2025short} introduces real entity names whose representations already carry semantic information from pretraining. These experiments test whether the behavior observed with neutral synthetic artifacts extends to more realistic shortcuts.

\subsection{Naturally Occurring Gender Correlations}
\label{sec:bias-in-bios}

We use Bias in Bios, a corpus of 397,000 biographies annotated with profession and binary gender. We construct binary tasks pairing a male-typical profession with a female-typical profession and resample two training environments such that $P(\text{gender}=\text{canonical}(y)\mid Y=y)$ equals $0.99$ and $0.80$, respectively. This naturally occurring correlation is fully inverted at test time. We add a label-noise rate of $\epsilon=0.25$ and train the same frozen DistilBERT classifier as in the controlled experiments. To examine whether semantic separability affects shortcut reliance, we compare the five semantically closest and the five most semantically distant eligible profession pairs. Each training environment contains at least 1,000 examples, and results are averaged over three seeds. Further details on environment construction, pair selection, and semantic-distance estimation are provided in Appendix~\ref{app:bib-details}.

\renewcommand{\ms}[2]{%
  \ensuremath{#1{\scriptstyle\,\pm\,#2}}%
}

\begin{table}[t]
\centering
\footnotesize
\setlength{\tabcolsep}{2.5pt}
\renewcommand{\arraystretch}{1.05}

\begin{tabularx}{\columnwidth}{
    @{}
    >{\raggedright\arraybackslash}X
    ccc
    @{}
}
\toprule
\textbf{Profession pair}
& \textbf{ERM}
& \textbf{IRM}
& \textbf{$\Delta$} \\
\midrule

\multicolumn{4}{c}{\emph{Semantically close pairs}} \\
\cmidrule(lr){1-4}

attorney/paralegal
& \ms{8.4}{0.6}
& \ms{47.1}{0.5}
& \textcolor{blue}{\ms{+38.8}{0.7}} \\

professor/psychologist
& \ms{22.5}{0.5}
& \ms{79.6}{1.4}
& \textcolor{blue}{\ms{+57.1}{1.3}} \\

chiropractor/psychologist
& \ms{33.9}{3.0}
& \ms{77.4}{1.4}
& \textcolor{blue}{\ms{+43.5}{4.4}} \\

chiropractor/nurse
& \ms{41.9}{2.0}
& \ms{73.6}{3.2}
& \textcolor{blue}{\ms{+31.7}{5.2}} \\

physician/nurse
& \ms{35.0}{0.5}
& \ms{83.6}{0.9}
& \textcolor{blue}{\ms{+48.6}{1.1}} \\

\midrule

\multicolumn{4}{c}{\emph{Semantically distant pairs}} \\
\cmidrule(lr){1-4}

dj/nurse
& \ms{47.1}{2.4}
& \ms{91.7}{1.4}
& \textcolor{blue}{\ms{+44.6}{1.1}} \\

rapper/psychologist
& \ms{39.8}{2.3}
& \ms{79.4}{4.5}
& \textcolor{blue}{\ms{+39.6}{3.7}} \\

dj/paralegal
& \ms{74.9}{1.2}
& \ms{81.5}{14.1}
& \textcolor{blue}{\ms{+6.6}{13.2}} \\

rapper/paralegal
& \ms{77.3}{3.0}
& \ms{86.9}{3.5}
& \textcolor{blue}{\ms{+9.6}{0.5}} \\

rapper/nurse
& \ms{52.7}{1.8}
& \ms{87.0}{2.8}
& \textcolor{blue}{\ms{+34.3}{4.3}} \\

\bottomrule
\end{tabularx}

\caption{
Bias in Bios OOD accuracy for semantically close and distant profession pairs.
Results are mean $\pm$ standard deviation over three seeds; values are percentages.
}
\label{tab:bib-results}
\end{table}

Across the ten profession pairs, IRM improves mean OOD accuracy from $43.35\%$ to $78.78\%$, corresponding to an average gain of $35.44$ percentage points. The improvement is positive for every evaluated pair. The results are consistent with greater shortcut reliance when professions are semantically similar. For the five semantically closest pairs, ERM and IRM reach average OOD accuracies of $28.3\%$ and $72.3\%$, respectively, yielding a $43.9$ percentage-point gain. For the five most semantically distant pairs, they reach $58.4\%$ and $85.3\%$, yielding a smaller $26.9$ percentage-point gain. As expected, when profession pairs are semantically close, profession-specific cues are less discriminative, making gender a comparatively easier and more attractive signal for separating the classes. For semantically distant pairs, other textual features already provide strong predictive cues, reducing the relative usefulness of gender. IRM's average advantage is therefore $17.0$ percentage points larger for semantically close pairs, primarily because ERM is less robust when profession-specific representations are harder to separate.

\subsection{Semantically Loaded Entity Shortcuts}
\label{sec:actorcorr}

We next evaluate ActorCorr, which introduces actor names into IMDB reviews as semantically loaded sentiment shortcuts. At a fixed injection frequency of 5\%, Morgan Freeman is associated with positive sentiment and Adam Sandler with negative sentiment. We construct two training environments with shortcut purities $\pi^{(e)}\in\{1.0,0.85\}$, where $\pi^{(e)}$ is the probability that the injected actor agrees with the sentiment label. Unlike the neutral tokens used in our controlled experiments, these names may carry associations inherited from pretraining. Each test review is evaluated under three conditions: \emph{clean}, , corresponding to the original unmodified review; \emph{correlated}, with an actor agreeing with the label; and \emph{anti-correlated}, with an actor associated with the opposite label.

We report the Anti-Correlated Accuracy Change (ACAC; \citet{eshuijs2025short}), defined as the signed
average of the class-specific accuracy difference between the original (clean) and anti-correlated conditions:
\begin{equation}
    \operatorname{ACAC} = \frac{1}{2} \sum_{y \in \{\mathrm{pos},\mathrm{neg}\}} \left( \operatorname{Acc}^{(y)}_{\mathrm{clean}} - \operatorname{Acc}^{(y)}_{\mathrm{anti}} \right).
\label{eq:actorcorr-acac}
\end{equation}
Higher positive values indicate greater sensitivity to misleading actor cues. Because ACAC is signed, class-specific effects may partially cancel (we analyze this in Appendix~\ref{app:actorcorr-details}).

\begin{table}[t]
\centering
\footnotesize
\setlength{\tabcolsep}{5pt}
\begin{tabular}{@{}lccc@{}}
\toprule
\textbf{Model} & $\boldsymbol{\lambda}$ & \textbf{OOD} $\uparrow$ & \textbf{ACAC} $\downarrow$ \\
\midrule
ERM & -- & \ms{0.739}{0.004} & \ms{0.186}{0.002} \\
IRM & 10 & {\bfseries\boldmath \ms{0.745}{0.007}} & \ms{0.146}{0.002} \\
IRM & 100 & \ms{0.704}{0.008} & {\bfseries\boldmath \ms{0.049}{0.002}} \\
\bottomrule
\end{tabular}
\caption{ActorCorr aggregate results over three seeds. IRM with $\lambda=10$ achieves the highest pooled OOD accuracy, whereas IRM with $\lambda=100$ achieves the lowest signed ACAC.}
\label{tab:actorcorr-main}
\end{table}

Table~\ref{tab:actorcorr-main} shows a clear trade-off between OOD performance and shortcut invariance. ERM achieves an OOD accuracy of \ms{0.739}{0.004} with an ACAC of \ms{0.186}{0.002}. With $\lambda=10$, IRMv1 slightly improves OOD accuracy to \ms{0.745}{0.007} while reducing ACAC to \ms{0.146}{0.002}. Increasing the penalty to $\lambda=100$ further lowers ACAC to \ms{0.049}{0.002}, but OOD accuracy drops to \ms{0.704}{0.008}. Thus, larger penalties reduce sensitivity to the spurious actor cue but can also hurt predictive performance, requiring a trade-off between robustness and OOD accuracy. The low ACAC at $\lambda=100$ should nevertheless be interpreted cautiously because of partial class-specific cancellation (Appendix~\ref{app:actorcorr-details}).

Table~\ref{tab:actorcorr-conditions} highlights the contrasting behavior of ERM and IRMv1 ($\lambda=100$) across evaluation conditions. ERM relies heavily on the actor cue: its accuracy rises to $0.870$ when the cue is correlated with the label but falls to $0.574$ when it becomes misleading. In contrast, IRMv1 is less sensitive to this shift, with accuracies of $0.726$ and $0.671$ in the correlated and anti-correlated conditions, respectively, reducing the condition spread from $0.296$ to $0.055$. This robustness comes at the cost of lower clean and correlated accuracy, illustrating how invariant learning can suppress a semantically meaningful shortcut while sacrificing some predictive performance. The full penalty sweep and class-specific results are reported in Appendix~\ref{app:actorcorr-details}.

\begin{table}[t]
\centering
\footnotesize
\setlength{\tabcolsep}{4pt}
\renewcommand{\arraystretch}{1.08}
\begin{tabular}{@{}lcc@{}}
\toprule
\textbf{Metric} & \textbf{ERM} & \textbf{IRM} \\
\midrule
ID accuracy & \ms{0.755}{0.002} & \ms{0.733}{0.003} \\
Pooled OOD accuracy & \ms{0.739}{0.004} & \ms{0.704}{0.008} \\
\midrule
Clean accuracy & 0.759 & 0.716 \\
Correlated accuracy & 0.870 & 0.726 \\
Anti-correlated accuracy & 0.574 & 0.671 \\
Spread $\downarrow$ & 0.296 & 0.055 \\
\midrule
ACAC $\downarrow$ & \ms{0.186}{0.002} & \ms{0.049}{0.002} \\
\bottomrule
\end{tabular}
\caption{ActorCorr condition-level results over three seeds for ERM and IRMv1 with $\lambda=100$. Spread is the difference between the highest and lowest condition-level accuracies.}
\label{tab:actorcorr-conditions}
\end{table}

\section{Conclusion}

We systematically evaluated IRM across label-generated features, confounding, and selection bias. Under controlled shortcut shifts, IRM consistently improves OOD robustness across causal tasks and succeeds in anti-causal settings with stable class prevalence. As the task-relevant signal weakens, IRM also remains more robust than ERM by favoring invariant predictors. Beyond controlled shortcuts, IRM improves OOD accuracy across all Bias in Bios profession pairs, with larger average gains for semantically similar professions, although semantic distance does not fully explain shortcut reliance. On ActorCorr, IRM reduces sensitivity to semantically loaded actor cues, but reveals an accuracy--invariance trade-off. Overall, IRM's effectiveness depends less on the shortcut's structural origin than on whether training environments expose its instability while preserving a stable task-relevant mechanism. These findings also show the importance of targeted interventions and class-specific metrics, as pooled OOD accuracy may conceal shortcut reliance. Extending this analysis to naturally occurring environments and fully fine-tuned language models remains an important direction for future work.

\section*{Limitations}

A first limitation concerns the controlled nature of our spurious mechanisms. Selection bias differs fundamentally from the other mechanisms considered here: it arises at the observation stage through conditioning on a collider, whereas confounding and label-generated features can be part of the underlying data-generating process itself. To isolate these mechanisms while retaining real-world text semantics, we introduce the corresponding shortcuts post hoc through controlled token injection. This design enables precise manipulation of the relevant causal relationships, but does not capture spurious correlations that emerge organically in natural text, such as environment-dependent shifts in style, lexical choice, or discourse. Our injected cues should therefore be viewed as controlled proxies for more diffuse real-world shortcuts, whose causal sources are substantially harder to isolate. A second limitation concerns the scope of IRM theory. Although IRM has been widely studied in classification settings \citep{arjovsky2019invariant, ahuja2020invariant, choe2020empirical, kamath2021does}, its original theoretical guarantees were developed primarily for regression. Extending invariance principles to classification remains an active research problem: subsequent work provides guarantees under more specific assumptions \citep{ahuja2021invariance}, while their applicability to modern Transformer-based models remains less well understood \citep{peyrard2022invariant}. A third limitation is that we freeze the pretrained encoder and train only a linear classification head. This choice deliberately places ERM and IRMv1 in the same fixed representation space, allowing us to isolate the effect of the optimization objective, but it does not reflect the fully fine-tuned setting commonly used with language models. This is particularly relevant because IRM can behave differently in highly overparameterized models, where additional capacity may permit solutions that satisfy the invariance penalty without yielding the intended OOD generalization benefits \citep{rosenfeld2020risks, gulrajani2020search}. Extending the present analysis to fully fine-tuned Transformers therefore remains an important direction for future work. Finally, our study is designed to compare causal shortcut mechanisms under controlled interventions rather than to benchmark IRM against the broader bias-mitigation literature. Evaluating the same mechanisms in less controlled real-world settings would enable direct comparisons with representative alternatives such as Group DRO \citep{sagawa2019distributionally}, JTT \citep{liu2021just}, Learning from Failure (LfF) \citep{nam2020learning}, and bias-model-based reweighting \citep{utama2020towards}. Such comparisons would help clarify when the explicit environment supervision used by IRM provides advantages over methods based on group annotations or on identifying biased examples from model behavior.

\section*{Ethical Considerations}

We exclusively utilize publicly available benchmarks in strict accordance with their original licenses for research purposes. Our methodology involves the artificial synthesis of environments through token injection; this process does not involve the collection of new personally identifiable information or the use of private user data. To minimize the environmental footprint of our extensive sensitivity analyses, we conducted all experiments using DistilBERT \citep{sanh2019distilbert}, a smaller and more energy-efficient architecture than standard large-scale models.

\bibliography{references}

\appendix

\section{Model Architecture and Experimental Setup}
\label{app:model-setup}

Before detailing our environmental synthesis framework, we formally define the predictive architecture shared by ERM and IRMv1 and clarify its correspondence with the notation introduced in Section~\ref{sec:irm}. Let $\mathcal{X}$ denote the input text space and $\mathcal{Y}$ the downstream label space.

\paragraph{Encoder Architecture and Text Representation.} All experiments use a frozen DistilBERT encoder $\Phi_{\text{BERT}}: \mathcal{X} \rightarrow \mathbb{R}^{768}$. Specifically, we use DistilBERT (\texttt{distilbert-base-uncased}) \citep{sanh2019distilbert}, a knowledge-distilled variant of BERT-base with 6 Transformer layers, 12 attention heads, and approximately 66 million parameters. DistilBERT retains 97\% of BERT's language understanding capabilities while being 40\% smaller and 60\% faster, making it well suited to the extensive sensitivity analyses conducted in this study. Texts are tokenized using the DistilBERT WordPiece tokenizer with a maximum sequence length of 256 tokens for AG News \citep{zhang2015characterlevel}, and 512 tokens for IMDB Genres and Amazon Books. For any input sequence $x \in \mathcal{X}$, the encoder produces a fixed sentence-level representation $z(x) = \Phi_{\text{BERT}}(x) \in \mathbb{R}^{768}$, obtained by masked mean pooling over the final hidden states of all non-padding tokens. The parameters of $\Phi_{\text{BERT}}$ remain frozen throughout all experiments.

\paragraph{Classification Head and IRMv1 Parameterization.} A single trainable linear head $W_\theta \in \mathbb{R}^{d_{\text{out}} \times 768}$ is placed on top of the frozen representation, with $d_{\text{out}}=1$ for the binary Amazon Books and IMDB Genres tasks and $d_{\text{out}}=4$ for AG News. The resulting predictive logits are $\Phi_\theta(x) = W_\theta \Phi_{\text{BERT}}(x)$. We use $\Phi_\theta$ deliberately to match the notation of IRMv1. In Equation~\ref{eq:irmv1}, $\Phi$ denotes the parameterized mapping that directly outputs predictive logits. In our architecture, this mapping corresponds to the composition $W_\theta \circ \Phi_{\text{BERT}}$. Although $\Phi_{\text{BERT}}$ is fixed, $\Phi_\theta$ remains trainable through the parameters $\theta$ of $W_\theta$. Importantly, the matrix $W_\theta$ should not be confused with the scalar classifier $w$ appearing in the IRMv1 objective. The former is the actual trainable classification head, whereas the latter is the dummy scalar variable at which the invariance penalty is evaluated:
\begin{equation}
    \left\|
    \nabla_{w \mid w=1}
    R^e\!\left(w \cdot \Phi_\theta\right)
    \right\|^2.
\end{equation}
The dummy classifier $w$ is not an independently optimized model component; gradients of this penalty are propagated through $\Phi_\theta$ and therefore update $W_\theta$. For the binary Amazon Books and IMDB Genres tasks, $\Phi_\theta(x) \in \mathbb{R}$ and we use binary cross-entropy (BCE) with logits, with predictions obtained by thresholding the logit at zero. For AG News, $\Phi_\theta(x) \in \mathbb{R}^{4}$ and we use standard multiclass cross-entropy loss.

\paragraph{Optimization and Hardware.} Both ERM and IRM heads are trained using the Adam optimizer \citep{kingma2014adam} with hyperparameters $\beta_1=0.9$, $\beta_2=0.999$, $\epsilon=10^{-8}$, and a weight decay of $10^{-4}$. We use a batch size of 256. Each model is trained for 25,000 gradient steps. The learning rate and IRM penalty weight $\lambda$ are kept identical for both ERM and IRM within each experimental configuration to rigorously isolate the effect of the objective function. All reported results are averaged over three random seeds (0, 1, 2). Finally, given the compact size of our frozen architecture, all experiments were conducted efficiently using a single NVIDIA RTX A5000 GPU with 24~GB of VRAM.

\begin{figure*}[t]
\centering
\includegraphics[width=0.92\linewidth]{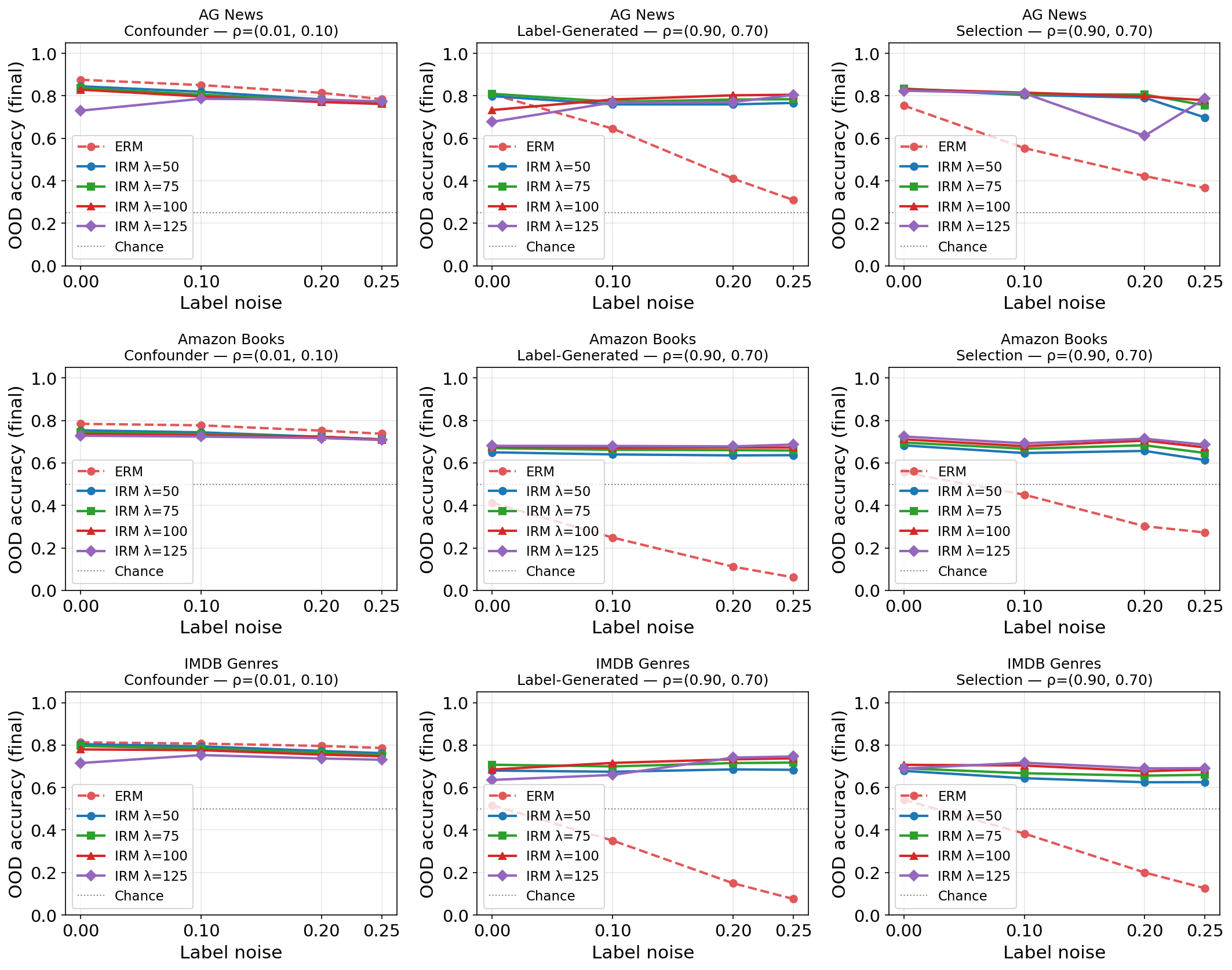}
\caption{OOD accuracy as a function of label noise, for each $\lambda \in \{50, 75, 100, 125\}$ (solid colored lines) and ERM (dashed red). Each panel corresponds to one dataset--mechanism pair. All IRM variants consistently outperform ERM. Performance differences across $\lambda$ values are modest, confirming robustness of the chosen $\lambda=100$.}
\label{fig:lambda-sensitivity}
\end{figure*}

\paragraph{Hyperparameter Selection: IRM Penalty Weight $\lambda$.} The penalty coefficient $\lambda$ in IRMv1 (Equation~\ref{eq:irmv1}) was selected via a grid search over $\lambda \in \{50, 75, 100, 125\}$, covering all dataset-mechanism combinations (3 datasets $\times$ 3 mechanisms $\times$ 4 noise levels $\times$ 2 correlation strengths, yielding 48 cells), with all other conditions held fixed (same seed, 25,000 gradient steps, learning rate $10^{-4}$). Figure~\ref{fig:lambda-sensitivity} shows the corresponding accuracy curves as a function of label noise, while Table~\ref{tab:hparam_search} reports the underlying OOD accuracies. All IRM variants consistently outperform ERM, and performance is broadly robust to the exact choice of $\lambda$; $\lambda=100$ achieves the best or near-best OOD accuracy in 31 out of 48 cells and is therefore fixed as the default for all main-paper experiments.

\begin{table*}[t]
\centering

\scriptsize
\setlength{\tabcolsep}{2.5pt}
\renewcommand{\arraystretch}{1.05}

\begin{tabular*}{\textwidth}{
@{\extracolsep{\fill}}
ll
cccc
cccc
cccc
@{}
}
\toprule
& &
\multicolumn{4}{c}{\textbf{AG News}} &
\multicolumn{4}{c}{\textbf{Amazon Books}} &
\multicolumn{4}{c}{\textbf{IMDB Genres}} \\
\cmidrule(lr){3-6}
\cmidrule(lr){7-10}
\cmidrule(lr){11-14}

\textbf{Mechanism} &
\textbf{Model} &
\multicolumn{4}{c}{\textbf{Label-noise rate } $\boldsymbol{\epsilon}$} &
\multicolumn{4}{c}{\textbf{Label-noise rate } $\boldsymbol{\epsilon}$} &
\multicolumn{4}{c}{\textbf{Label-noise rate } $\boldsymbol{\epsilon}$} \\

& &
\textbf{0.00} & \textbf{0.10} & \textbf{0.20} & \textbf{0.25} &
\textbf{0.00} & \textbf{0.10} & \textbf{0.20} & \textbf{0.25} &
\textbf{0.00} & \textbf{0.10} & \textbf{0.20} & \textbf{0.25} \\
\midrule

\multirow{5}{*}{Confounding}
& ERM
& 0.875 & 0.851 & 0.814 & 0.784
& 0.784 & 0.777 & 0.752 & 0.738
& 0.813 & 0.807 & 0.796 & 0.786 \\

& IRM ($\lambda=50$)
& \textbf{0.845} & \textbf{0.819} & \textbf{0.783} & 0.765
& \textbf{0.753} & \textbf{0.744} & \textbf{0.723} & 0.710
& \textbf{0.804} & \textbf{0.794} & \textbf{0.772} & \textbf{0.762} \\

& IRM ($\lambda=75$)
& 0.837 & 0.805 & 0.778 & 0.763
& 0.744 & 0.736 & 0.723 & \textbf{0.711}
& 0.795 & 0.784 & 0.763 & 0.753 \\

& IRM ($\lambda=100$)
& 0.829 & 0.797 & 0.770 & 0.761
& 0.737 & 0.730 & 0.723 & 0.710
& 0.779 & 0.776 & 0.755 & 0.748 \\

& IRM ($\lambda=125$)
& 0.730 & 0.786 & 0.781 & \textbf{0.774}
& 0.728 & 0.724 & 0.717 & 0.708
& 0.716 & 0.753 & 0.737 & 0.731 \\

\midrule

\multirow{5}{*}{Label-Generated}
& ERM
& 0.807 & 0.645 & 0.410 & 0.309
& 0.412 & 0.248 & 0.112 & 0.061
& 0.516 & 0.350 & 0.149 & 0.075 \\

& IRM ($\lambda=50$)
& 0.798 & 0.759 & 0.759 & 0.766
& 0.650 & 0.640 & 0.635 & 0.636
& 0.680 & 0.674 & 0.685 & 0.683 \\

& IRM ($\lambda=75$)
& \textbf{0.809} & 0.772 & 0.782 & 0.784
& 0.669 & 0.662 & 0.660 & 0.658
& \textbf{0.708} & 0.700 & 0.716 & 0.718 \\

& IRM ($\lambda=100$)
& 0.733 & \textbf{0.782} & \textbf{0.802} & \textbf{0.805}
& 0.676 & 0.671 & 0.672 & 0.672
& 0.685 & \textbf{0.716} & 0.733 & 0.738 \\

& IRM ($\lambda=125$)
& 0.677 & 0.768 & 0.771 & 0.803
& \textbf{0.680} & \textbf{0.680} & \textbf{0.678} & \textbf{0.686}
& 0.636 & 0.659 & \textbf{0.741} & \textbf{0.747} \\

\midrule

\multirow{5}{*}{Selection}
& ERM
& 0.754 & 0.554 & 0.422 & 0.366
& 0.555 & 0.451 & 0.302 & 0.273
& 0.542 & 0.384 & 0.200 & 0.126 \\

& IRM ($\lambda=50$)
& 0.830 & 0.804 & 0.791 & 0.698
& 0.682 & 0.646 & 0.656 & 0.614
& 0.678 & 0.644 & 0.625 & 0.625 \\

& IRM ($\lambda=75$)
& \textbf{0.834} & 0.807 & \textbf{0.806} & 0.755
& 0.696 & 0.667 & 0.683 & 0.647
& 0.691 & 0.667 & 0.656 & 0.660 \\

& IRM ($\lambda=100$)
& 0.831 & \textbf{0.815} & 0.797 & 0.779
& 0.711 & 0.679 & 0.705 & 0.673
& \textbf{0.707} & 0.704 & 0.677 & 0.685 \\

& IRM ($\lambda=125$)
& 0.824 & 0.812 & 0.612 & \textbf{0.786}
& \textbf{0.725} & \textbf{0.693} & \textbf{0.714} & \textbf{0.686}
& 0.690 & \textbf{0.717} & \textbf{0.691} & \textbf{0.692} \\

\bottomrule
\end{tabular*}
\caption{
Hyperparameter search results. Mean final OOD accuracy for IRM penalty
values $\lambda\in\{50,75,100,125\}$ across label-noise rates
$\epsilon\in\{0.00,0.10,0.20,0.25\}$. Results are averaged over
spurious-correlation levels. Bold indicates the best IRM result for
each dataset, mechanism, and noise configuration.
}
\label{tab:hparam_search}
\end{table*}

\section{Structural Equation Models and the Nature of Environments}
\label{app:sem}

In IRM theory, the algorithm exploits natural datasets collected from distinct temporal or spatial contexts to identify invariant correlations. However, since we synthesize our own environments by injecting misleading correlations, it is imperative to demonstrate the scientific validity of this procedure. This justification relies on the formal framework of Structural Equation Models (SEM).

\begin{definition}[SEM]
An SEM $\mathcal{C} := (\mathcal{S}, N)$ governing a random vector $X = (X_1, \ldots, X_d)$ is a set of structural equations:
\[
\mathcal{S}_i: X_i \leftarrow f_i(\operatorname{Pa}(X_i), N_i)
\]
where $\operatorname{Pa}(X_i) \subseteq \{X_1, \ldots, X_d\} \setminus \{X_i\}$ denotes the set of causal parents of $X_i$, and $N_i$ are independent noise variables. We say that $X_i$ causes $X_j$ if $X_i \in \operatorname{Pa}(X_j)$. The causal graph of $X$ is obtained by drawing an edge from $X_i$ to $X_j$ if $X_i \in \operatorname{Pa}(X_j)$, assuming the graph is a Directed Acyclic Graph (DAG).
\end{definition}

Within this framework, an environment $e$ is defined as the result of an intervention on the SEM \citep{peters2016causal, arjovsky2019invariant}, consisting of replacing one or more of its structural equations to obtain a new SEM $\mathcal{C}^e = (\mathcal{S}^e, N^e)$. For IRM's guarantees to be theoretically legitimate, these interventions must be ``valid'', satisfying three criteria:
\begin{itemize}
\item The causal graph remains acyclic.
\item The conditional expectation remains invariant: $\mathbb{E}[Y^e \mid \operatorname{Pa}(Y)] = \mathbb{E}[Y \mid \operatorname{Pa}(Y)]$.
\item The variance $\mathbb{V}[Y^e \mid \operatorname{Pa}(Y)]$ remains within a finite range.
\end{itemize}

Our protocol strictly respects these conditions. By injecting spurious variables without altering the intrinsic semantic link between the text and its label, we generate our training environments $\mathcal{E}_{\text{tr}}$ via valid interventions by mathematically modulating the strength of these spurious correlations. This approach ensures that any observed failure of IRM stems from a structural limitation of the tested mechanism rather than a violation of the method's core assumptions. This structural rigor is fundamental to the analysis in the following section, where we evaluate confounding. By distinguishing between three possible modeling scenarios, we clarify which interventions preserve the target mechanism and why invariance of the causal mechanism need not imply invariance of the optimal predictor when relevant causal parents are latent.

\section{Confounding and IRM Validity}
\label{app:confounding-validity}

Unlike the other two processes, the confounder $U$ exerts a causal influence on the label $Y$. For IRM to be theoretically valid, the influence of causal parents on the target must remain stable, as established in Appendix~\ref{app:sem}. We distinguish three environment-generation strategies. The first two illustrate, respectively, a valid intervention on the shortcut mechanism and an invalid intervention on the target mechanism. The third is also structurally valid under the criterion of Appendix~\ref{app:sem}, but highlights the distinction between invariance of the causal mechanism and invariance of the optimal predictor when causal parents are latent.

\begin{itemize}
    \item \textbf{Varying proxy:} The link between the unobserved confounder $U$ and the latent concept $Z$ varies across environments, but the $U \rightarrow Y$ relationship remains fixed. This is the intervention used in our main confounding experiments: the causal rule remains fixed, and only the misleading proxy relationship fluctuates. This setup preserves the condition $\mathbb{E}[Y^e \mid \operatorname{Pa}(Y)] = \mathbb{E}[Y \mid \operatorname{Pa}(Y)]$.
    \item \textbf{Varying causal strength:} In this scenario, the direct causal link between the confounder $U$ and the target $Y$ varies. This represents a direct violation of IRM's assumptions, as the conditional expectation $\mathbb{E}[Y \mid \operatorname{Pa}(Y)]$ is no longer invariant. Since no stable relationship exists across environments, IRM systematically fails to identify a robust predictor.
    \item \textbf{Confounder-prevalence shift:} A third possibility is to keep both structural mechanisms \(P(Y\mid U)\) and \(P(Z\mid U)\) fixed while changing the marginal distribution \(P^{(e)}(U)\) across environments. Unlike a change in the \(U\rightarrow Y\) mechanism, such a prevalence shift does not by itself violate the validity condition introduced in Appendix~\ref{app:sem}: the causal relationship between \(Y\) and its parent \(U\) remains unchanged, and therefore \(\mathbb{E}[Y^{e}\mid U]=\mathbb{E}[Y\mid U]\). However, because \(U\) is unobserved, changing its prevalence can alter predictive conditionals involving observed proxies. In particular, $P^{(e)}(Y\mid Z) = \sum_u P(Y\mid u)P^{(e)}(u\mid Z)$, and \(P^{(e)}(u\mid Z)\) generally depends on \(P^{(e)}(U)\). Consequently, a prevalence shift may change the optimal predictor based on the observed representation even though the underlying causal mechanism remains invariant. Whether IRM can recover an invariant predictor then depends on whether the observed representation contains sufficient stable information about the target independently of the prevalence-sensitive proxy.
\end{itemize}

\section{The Role of Causal Direction}
\label{app:causal-direction}

IRM seeks a representation for which the same predictor remains optimal across environments. However, its formal guarantees rely on a connection between invariance and causality \citep{arjovsky2020outof, arjovsky2019invariant}. The implications of this connection depend on the generative direction of the prediction task: causal learning ($X\rightarrow Y$) or anti-causal learning ($Y\rightarrow X$). This distinction can be understood through the Principle of Independent Causal Mechanisms (ICM) \citep{scholkopf2012causal, peters2017elements}. The ICM principle states that a causal generative process consists of autonomous mechanisms: the conditional distribution of an effect given its cause is structurally independent of the marginal distribution of that cause.

\subsection{Causal Learning}

In a causal prediction task ($X\rightarrow Y$), such as topic classification or review-helpfulness prediction, the observed features $X$ causally determine the target $Y$. Under the ICM principle, the mechanism $P(Y\mid X)$ may remain stable even when the distribution of inputs $P^{(e)}(X)$ changes across environments. The joint distribution in environment $e$ can be factorized as
\begin{equation}
P^{(e)}(X,Y) = P^{(e)}(X)P(Y\mid X),
\label{eq:causal-factorization}
\end{equation}
where environmental variation affects $P^{(e)}(X)$ but leaves the causal mechanism $P(Y\mid X)$ unchanged. In this setting, identifying an invariant representation amounts to preserving the direct causal parents of $Y$, which define a stable and optimal predictor across valid interventions \citep{arjovsky2020outof, arjovsky2019invariant}. IRM is therefore theoretically well aligned with causal prediction when the environmental changes do not modify the mechanism generating $Y$ from its causes.

\subsection{Anti-Causal Learning}

In an anti-causal prediction task ($Y\rightarrow X$), the target variable generates the observed features. Sentiment classification provides a typical example: the latent sentiment $Y$ generates the vocabulary, expressions, and other textual features contained in $X$. Predicting $Y$ from $X$ therefore amounts to inferring a cause from its effects \citep{jin-etal-2021-causal, feder2022causal}. Under the ICM principle, the stable mechanism is now the generative distribution $P(X\mid Y)$ rather than the predictive distribution $P(Y\mid X)$. The environment-specific joint distribution factorizes as
\begin{equation}
P^{(e)}(X,Y) = P(X\mid Y)P^{(e)}(Y),
\label{eq:anticausal-factorization}
\end{equation}
where $P(X\mid Y)$ is assumed to remain stable, but the class prevalence $P^{(e)}(Y)$ may vary across environments. A classifier estimates the posterior distribution $P^{(e)}(Y\mid X)$, which follows from Bayes' theorem:
\begin{equation}
\begin{aligned}
P^{(e)}(Y \mid X) &= \frac{P(X \mid Y)P^{(e)}(Y)} {P^{(e)}(X)} \\[3pt] &= \frac{P(X \mid Y)P^{(e)}(Y)} {\displaystyle \sum_{y' \in \mathcal{Y}} P(X \mid Y=y')P^{(e)}(Y=y')}
\end{aligned}
\label{eq:bayes}
\end{equation}

where $\mathcal{Y}$ denotes the set of possible target labels. Even if the generative mechanism $P(X\mid Y)$ remains invariant, the optimal predictive distribution $P^{(e)}(Y\mid X)$ depends explicitly on the environment-specific class prevalence $P^{(e)}(Y)$. This creates two distinct regimes for invariant learning:
\begin{itemize}
    \item \textbf{Stable class prevalence.} If $P^{(e)}(Y)$ is identical across training and test environments, and $P(X\mid Y)$ remains stable, then $P^{(e)}(Y\mid X)$ is also invariant. IRM can suppress unstable environmental shortcuts while retaining predictive relationships shared across environments.
    \item \textbf{Label shift.} If $P^{(e)}(Y)$ changes across environments, then the optimal posterior $P^{(e)}(Y\mid X)$ generally changes even though $P(X\mid Y)$ remains stable. Enforcing a common optimal predictor across environments then conflicts with the environment-specific predictor required by Bayes' rule.
\end{itemize}

Label shift therefore creates a structural difficulty for IRM in anti-causal tasks. The problem is not that the underlying generative mechanism becomes unstable, but that the posterior used for prediction depends on a changing class prior. A strict invariance constraint may consequently prevent the classifier from adapting to the posterior required in each environment.

\subsection{Implications for Our Experiments}

Our anti-causal experiments are conducted with balanced and constant class proportions across training and test environments. This construction keeps $P^{(e)}(Y)$ fixed while varying only the association between the target and the designated shortcut. Under the assumed stability of $P(X\mid Y)$, the optimal predictive mechanism can therefore remain invariant across environments. This design isolates IRM's ability to reject unstable shortcuts without simultaneously introducing label shift. The resulting performance should consequently be interpreted as evidence that IRM can operate in an anti-causal setting under stable class prevalence, rather than as evidence that it is generally robust to arbitrary distribution shifts in anti-causal NLP tasks.

\section{Methodological Justification of Sequence Length as a Spurious Feature}
\label{app:seqlen}

To ensure that our experimental design in the IMDB Genres setting is theoretically justified, it is necessary to establish that text description length is a feature that models can readily extract and exploit as a predictive shortcut. Linguistic probing and model analysis literature establish that sequence length information is preserved within sentence-level vector spaces \citep{adi2016finegrained}. Consequently, when sentence representations are computed via token averaging (mean pooling), the resulting vector is directly bound to the sequence length $N$, as embedding norms predictably decrease as sentences grow longer due to the properties of natural language distributions. Transformer-based architectures are highly susceptible to the sequence length learning problem whenever downstream classes exhibit asymmetrical length distributions \citep{baillargeon2022assessing}. In these unbalanced scenarios, standard ERM objectives tend to opportunistically exploit sequence length as a primary predictive feature instead of relying on genuine textual or semantic information. Furthermore, because sequence length distributions effectively behave like a distinct domain or environmental property during training, a significant mismatch or inversion of these length patterns at test time routinely triggers an overfitting collapse \citep{varis-bojar-2021-sequence}. Modulating this engineered length bias across training environments therefore provides a theoretically justified and rigorous proxy for evaluating the capacity of IRM to bypass surface heuristics in favor of invariant features.

\section{Detailed Construction of Spurious Mechanisms}
\label{app:mechanisms}

To rigorously evaluate IRM across diverse causal structures, we artificially inject or induce spurious correlations in each of the three datasets. Each corpus is partitioned into an 80\% training set, a 10\% in-distribution validation set, and a 10\% out-of-distribution test set. The training partition is further divided equally into two environments $\mathcal{E}_{\text{tr}} = \{e_1, e_2\}$, characterized by contrasting spurious correlation strengths, so as to create the distributional instability required by the invariance penalty. We detail below the exact construction procedure for each of the three mechanisms.

\paragraph{Label-Generated Features ($p_c$).} The target label $Y$ causally generates a discrete textual artifact, which is embedded in the raw text prior to encoding. For each dataset, we associate a set of semantically neutral vocabulary tokens to the possible classes: \{\texttt{moon}, \texttt{sun}\} for Amazon Books, \{\texttt{red}, \texttt{blue}, \texttt{green}, \texttt{yellow}\} for AG News (one per class), and \{\texttt{pine}, \texttt{ash}\} for IMDB Genres. Rather than simply prepending a single token at the start of the text, the selected token is inserted immediately before every occurrence of a predefined list of high-frequency, content-free function words (e.g., \emph{the, a, of, in, to}). This distributes the spurious signal across the entire sequence, making the shortcut substantially harder for ERM to ignore.

The correlation strength is controlled by $p_c = P(Z=y \mid Y=y)$. To prevent the true semantic signal from trivially dominating, a uniform label noise $\epsilon = 0.2$ is applied globally to the entire training partition before splitting into environments, ensuring that both environments share the same noise pattern. The two training environments are then constructed by injecting tokens with $p_c \in \{0.70, 0.90\}$ respectively. In the OOD test set, this correlation is completely inverted ($p_c = 0.0$), so the injected token systematically identifies the wrong class.

\paragraph{Confounding ($a$).} An unobserved confounder $C$ introduces a spurious backdoor path to the observed label. For each training example, $C$ is drawn uniformly from all possible classes for a fraction $p_{c\text{-flip}} = 0.50$ of the data; for the remainder, $C$ coincides with the true label $Y^*$ (inactive confounder). The observed label $Y_{\text{obs}}$ is then biased toward $C$: when $Y^* \neq C$, it is replaced by $C$ with probability $\gamma$ (set to 0.8 for Amazon and AG News, 0.5 for IMDB Genres). The model never observes $C$ directly; instead, it observes a noisy proxy $Z$, generated as $Z^{(e)} = C \oplus \operatorname{Ber}(a^{(e)})$, where $\oplus$ is the XOR operator and $a^{(e)}$ governs how faithfully $Z$ tracks $C$. The injected token encodes $Z$ rather than $Y_{\text{obs}}$, so the spurious correlation propagates through the backdoor path $Y_{\text{obs}} \leftarrow C \rightarrow Z \rightarrow X_Y^{\perp}$. The two training environments differ solely in the proxy noise rate: $a^{(e)} \in \{0.01, 0.10\}$, keeping $Z$ a reliable proxy in both environments. In the OOD test set, $a$ is set to 1.0, making $Z$ perfectly anti-correlated with $C$; evaluation is performed on the clean labels $Y^*$ (without label bias), maximally penalizing any model that has relied on the proxy token.

\paragraph{Selection Bias ($p_s$).} Unlike the token injection mechanisms above, selection bias operates purely by subsampling the corpus based on an unobserved collider $S$. A pre-existing latent feature $Z$ is defined for each dataset, and a sample is included in training ($S=1$) based on whether its $(Y,Z)$ pair is typical (aligned) or atypical (misaligned). The parameter $p_s$ controls the proportion of typical examples retained in each environment ($p_s \in \{0.70, 0.90\}$). The OOD test set contains exclusively atypical examples, completely inverting the spurious alignment.

For Amazon Books, the latent feature $Z$ is derived from the star rating: reviews with 4--5 stars are considered positive ($Z=1$) and reviews with 1--2 stars negative ($Z=0$). A typical example is one where $Z$ aligns with $Y$: helpful reviews are positive, unhelpful reviews are negative. The OOD set consists exclusively of counter-heuristic instances (helpful reviews with negative sentiment, and vice versa).

For AG News, $Z$ is the character length of the article, discretized into four global length bins via corpus-wide quartiles (Q1: $\leq P_{25}$, Q2: $P_{25}$--$P_{50}$, Q3: $P_{50}$--$P_{75}$, Q4: $> P_{75}$). Each class is assigned a dedicated typical bin: World $\rightarrow$ Q1, Sports $\rightarrow$ Q2, Business $\rightarrow$ Q3, Sci/Tech $\rightarrow$ Q4. In the OOD test set, each class is evaluated exclusively on articles from its most distant bin (i.e., Q1 $\leftrightarrow$ Q4 and Q2 $\leftrightarrow$ Q4 or Q1), ensuring the length cue is maximally misleading at test time.

For IMDB Genres, $Z$ is the description length, thresholded at the global P25 and P75 percentiles to obtain a short ($< P_{25}$) and a long ($> P_{75}$) category; intermediate-length texts are excluded. A typical example is a long Romance or a short Thriller. Training environments are assembled from $p_s \in \{0.70, 0.90\}$ typical pairs, while the OOD set contains exclusively atypical instances: short Romances and long Thrillers.

\section{Additional Bias in Bios Details}
\label{app:bib-details}

This section provides supplementary details on the construction of the Bias in Bios experiments introduced in Section~\ref{sec:bias-in-bios}, including the environment-generation procedure, the semantic-distance analysis used to select profession pairs, and an additional multiclass experiment.

\subsection{Dataset and Environment Construction} Bias in Bios contains approximately 400,000 short biographies annotated with one of 28 professions and binary gender \citep{de-arteaga2019bias}. We construct binary classification tasks from pairs containing one male-typical and one female-typical profession. A profession is defined as male-typical when fewer than 50\% of its biographies are annotated as female, and female-typical otherwise. Let $Y$ denote the profession label, $G$ the gender annotation, and $\operatorname{canonical}(Y)$ the majority gender of profession $Y$ in the original corpus. Shortcut alignment is defined as
\begin{equation}
    A = \mathbf{1}\!\left[G=\operatorname{canonical}(Y)\right].
\label{eq:bib-alignment}
\end{equation}
Gender is not provided as a separate input, but may be inferred from names, pronouns, and other gender-associated expressions. Two training environments are produced by resampling biographies within each profession while preserving class balance. Their shortcut strengths are
\begin{equation}
\begin{aligned}
P^{(e_1)}(A=1\mid Y)=0.99, \\
P^{(e_2)}(A=1\mid Y)=0.80.
\end{aligned}
\label{eq:bib-training-environments}
\end{equation}
The OOD test environment fully reverses the association, $P^{(\mathrm{test})}(A=1\mid Y)=0$, so that every test biography belongs to the minority-gender subset of its profession. A classifier relying on the training-time gender association therefore receives a systematically misleading cue at test time. Because profession-specific vocabulary is highly predictive, we apply a label-flip rate of $\epsilon=0.25$ to the training labels to make the gender cue competitive. Gender annotations and biography texts are left unchanged, and evaluation uses the original profession labels. We use the frozen DistilBERT encoder and masked mean pooling described in Appendix~\ref{app:model-setup}. Only the linear classification head is optimized. ERM pools the two environments, whereas IRMv1 retains the environment partition and uses a penalty coefficient of $\lambda=1000$. Both models are trained for 3,000 steps with Adam and a learning rate of $10^{-4}$.

\subsection{Semantic Distance and Pair Selection}
\label{app:bib-semantic-distance}

We quantify profession similarity in the representation space of the same frozen encoder. For each profession $c$, up to 300 biographies are sampled from the full corpus and encoded using masked mean pooling. The normalized profession centroid is
\begin{equation}
\tilde{\phi}_c = \frac{ \frac{1}{n_c}\sum_{i:Y_i=c}\Phi_{\mathrm{BERT}}(x_i) }{ \left\| \frac{1}{n_c}\sum_{i:Y_i=c}\Phi_{\mathrm{BERT}}(x_i) \right\|_2 }.
\label{eq:bib-centroid}
\end{equation}
For a male-typical profession $c_m$ and a female-typical profession $c_f$, their distance is
\begin{equation}
d(c_m,c_f) = 1-\tilde{\phi}_{c_m}^{\top}\tilde{\phi}_{c_f}.
\label{eq:bib-cosine-distance}
\end{equation}
After applying the corpus-availability criteria used in preprocessing, 160 cross-gender profession pairs are retained and ranked by this distance. Lower values indicate greater similarity in the frozen representation space. For the experiments in Section~\ref{sec:bias-in-bios}, we select five semantically close and five semantically distant eligible pairs while requiring enough examples for repeated balanced resampling with at least 1,000 examples per training environment. The ten evaluated pairs and their OOD results are reported in Table~\ref{tab:bib-results}. The close group is \professionpair{attorney}{paralegal}, \professionpair{professor}{psychologist}, \professionpair{chiropractor}{psychologist}, \professionpair{chiropractor}{nurse}, and \professionpair{physician}{nurse}; the distant group is \professionpair{dj}{nurse}, \professionpair{rapper}{psychologist}, \professionpair{dj}{paralegal}, \professionpair{rapper}{paralegal}, and \professionpair{rapper}{nurse}.

Semantic distance is used only as a descriptive proxy for the availability of profession-specific information. It does not directly measure the accessibility of gender, the frequency of explicit gender markers, within-class dispersion, or the orientation of the learned decision boundary. Consequently, distance should not be expected to predict ERM shortcut reliance or the IRM gain monotonically. Its role is to construct comparisons with different levels of representation-space separability, rather than to provide a complete model of shortcut attractiveness. For reference, the closest pair in the retained ranking is \professionpair{attorney}{paralegal}, with distance $0.0094$, while the maximum observed distance is $0.1489$ for \professionpair{rapper}{nurse}. The complete ranking is provided with the accompanying experimental material.

\subsection{Supplementary Multiclass Experiment}
\label{app:bib-multiclass}

We additionally construct a six-class task to test whether the shortcut remains competitive when the prediction space contains professions from several distinct domains. Training environments are created using the same resampling principle, and the gender--profession association is inverted at test time. In this single-seed experiment, ERM obtains an ID validation accuracy of $0.897$ and an OOD accuracy of $0.706$, compared with a chance level of $1/6$. Only $25.6\%$ of its predictions follow the inverted gender cue. These results suggest that gender is less competitive when profession-specific content provides strong class separation. This experiment remains supplementary because it uses a single seed and is not directly comparable to the binary tasks reported in Table~\ref{tab:bib-results}.

\subsubsection{Model and Optimization}
\label{app:bib-optimization}

We use the same frozen DistilBERT encoder and linear classification head described in Appendix~\ref{app:model-setup}. Biography representations are obtained using masked mean pooling over the final hidden states. Only the linear classification head is optimized. ERM pools the two training environments and minimizes the standard classification loss. IRMv1 retains the environment partition and optimizes the sum of the environment-specific risks and the invariance penalty. Both models are trained for 3,000 gradient steps with Adam and learning rate $10^{-4}$. For this experiment, the IRMv1 penalty coefficient is set to $\lambda=1000$. The larger penalty relative to the controlled experiments reflects the strong profession-specific semantic signal in Bias in Bios and the more diffuse nature of the demographic shortcut. All other model components are shared between ERM and IRM.

\section{ActorCorr: Experimental Details and Additional Results}
\label{app:actorcorr-details}

\subsection{Experimental Setup}
\label{app:actorcorr-setup}

ActorCorr is based on IMDB movie reviews and the actor-name shortcut protocol of \citet{eshuijs2025short}. We use the authors' data-generation pipeline, which identifies existing actor mentions through named-entity recognition and replaces an eligible mention at a fixed injection frequency of 5\%. Following the original pairing, Morgan Freeman is associated with positive sentiment and Adam Sandler with negative sentiment. Let $Z$ denote the injected actor and $Y$ the sentiment label. Shortcut purity in environment $e$ is defined as
\begin{equation} \pi^{(e)} = P^{(e)} \left(Z=z_{\mathrm{canonical}}(Y) \mid \text{actor injected}\right), \label{eq:actorcorr-purity}
\end{equation}
where $z_{\mathrm{canonical}}(Y)$ is the actor assigned to class $Y$. We construct two training environments with \begin{equation}
\pi^{(e_1)}=1.0, \qquad \pi^{(e_2)}=0.85.
\end{equation}

The injection frequency is identical across environments, and only the actor--label agreement varies. Unlike the neutral tokens used in the controlled experiments, actor names may carry associations inherited from pretraining. ActorCorr therefore tests IRM against a semantically loaded shortcut. We use the same frozen DistilBERT encoder and trainable linear classification head as in the controlled experiments. ERM pools both environments, whereas IRMv1 retains the environment partition when computing its risk and invariance penalty. Models are trained for 25,000 gradient steps, and results are averaged over three random seeds. The main ActorCorr experiment uses $\lambda=100$. Each test review is evaluated under three conditions:
\begin{itemize}
    \item \textbf{Clean}: no actor cue is injected;
    \item \textbf{Correlated}: the actor agrees with the sentiment label;
    \item \textbf{Anti-correlated}: the actor is associated with the opposite label.
\end{itemize}

To measure shortcut sensitivity independently of pooled accuracy, we compare each review with its anti-correlated variant, in which the injected actor cue points toward the incorrect sentiment class. For each sentiment class $y\in\mathcal{Y}$, we define the class-specific clean-to-anti-correlated accuracy gap as \begin{equation} G_y = \operatorname{Acc}_{\mathrm{clean}}^{(y)} - \operatorname{Acc}_{\mathrm{anti}}^{(y)}, \label{eq:actorcorr-class-gap} \end{equation} where $\operatorname{Acc}_{\mathrm{clean}}^{(y)}$ is the accuracy on clean reviews from class $y$, and $\operatorname{Acc}_{\mathrm{anti}}^{(y)}$ is the accuracy on the same class under the anti-correlated actor intervention. Since ActorCorr is a binary sentiment-classification task with $\mathcal{Y}=\{\mathrm{neg},\mathrm{pos}\}$, this definition yields two class-specific gaps: \begin{align} G_{\mathrm{neg}} &= \operatorname{Acc}_{\mathrm{clean}}^{(\mathrm{neg})} - \operatorname{Acc}_{\mathrm{anti}}^{(\mathrm{neg})}, \label{eq:actorcorr-gap-neg} \\ G_{\mathrm{pos}} &= \operatorname{Acc}_{\mathrm{clean}}^{(\mathrm{pos})} - \operatorname{Acc}_{\mathrm{anti}}^{(\mathrm{pos})}. \label{eq:actorcorr-gap-pos} \end{align} Here, $G_{\mathrm{neg}}$ measures the accuracy loss caused by injecting the positive-associated actor into negative reviews, whereas $G_{\mathrm{pos}}$ measures the loss caused by injecting the negative-associated actor into positive reviews. These gaps therefore quantify how strongly the misleading actor cue affects each sentiment class. A large positive value of $G_y$ indicates that the anti-correlated actor reduces accuracy, showing that the model remains sensitive to the shortcut for class $y$. A value close to zero indicates that the intervention has little effect. A negative value means that accuracy is higher under the anti-correlated intervention than in the clean condition, indicating a sign-inverted response rather than straightforward invariance. We summarize the model's average signed sensitivity using the Anti-Correlated Accuracy Change: \begin{equation} \operatorname{ACAC} = \frac{1}{|\mathcal{Y}|} \sum_{y\in\mathcal{Y}}G_y. \label{eq:actorcorr-acac-app} \end{equation} Because ActorCorr contains two sentiment classes, this expression reduces to \begin{equation} \operatorname{ACAC} = \frac{ G_{\mathrm{neg}}+G_{\mathrm{pos}} }{2}. \label{eq:actorcorr-acac-binary} \end{equation} A large positive ACAC indicates that anti-correlated actor cues reduce accuracy on average, suggesting reliance on the shortcut. An ACAC close to zero indicates limited average signed sensitivity. However, reporting $G_{\mathrm{neg}}$ and $G_{\mathrm{pos}}$ in addition to ACAC is important because the effect of the actor cue may be asymmetric across sentiment classes. In particular, a positive gap for one class and a negative gap for the other can partially cancel in the macro-average, producing a low ACAC even though the model remains sensitive to the intervention. Consequently, a robust model should ideally combine high clean and anti-correlated accuracy with both $G_{\mathrm{neg}}$ and $G_{\mathrm{pos}}$ close to zero. The aggregate ACAC measures average signed sensitivity, while the two class-specific gaps show whether that apparent invariance holds consistently across both sentiment classes.

\subsection{Results by Test Condition}
\label{app:actorcorr-variant-results}

Table~\ref{tab:actorcorr-variants} decomposes the pooled OOD results by test condition.
\begin{table}[t]
\centering
\footnotesize
\setlength{\tabcolsep}{4pt}
\renewcommand{\arraystretch}{1.12}
\begin{tabular}{@{}lcc@{}}
\toprule
\textbf{Test condition} & \textbf{ERM} & \textbf{IRM} \\
\midrule Clean & 0.759 & 0.716 \\ Correlated & 0.870 & 0.726 \\
Anti-correlated & 0.574 & 0.671 \\ \midrule Spread & 0.296 & 0.055 \\
\bottomrule
\end{tabular}
\caption{ActorCorr accuracy by test condition, averaged over three seeds. Spread denotes the difference between the highest and lowest condition-level accuracies.}
\label{tab:actorcorr-variants}
\end{table}

ERM strongly benefits from the correlated actor cue but deteriorates when the cue contradicts the label, producing a spread of $0.296$ across test conditions. IRM is less accurate on clean and correlated examples, but substantially more accurate under the anti-correlated intervention. Its spread is reduced to $0.055$, indicating more homogeneous behavior across conditions. Pooled OOD accuracy therefore understates the difference in shortcut reliance, since ERM's gains on correlated examples partially compensate for its failures on anti-correlated examples.

\subsection{Sensitivity to the Invariance Penalty}
\label{app:actorcorr-lambda}

Because ActorCorr combines a relatively small purity difference across environments with semantically meaningful actor names, we evaluate the following penalty values:
\begin{equation}
\lambda \in \{10,25,50,75,100\}.
\end{equation}
All configurations use the same architecture, training environments, and optimization protocol.
\begin{table}[t]
\centering
\scriptsize
\setlength{\tabcolsep}{1.5pt}
\renewcommand{\arraystretch}{1.08}
\resizebox{\columnwidth}{!}{
\begin{tabular}{@{}ccccc@{}}
\toprule $\lambda$ & \textbf{OOD} & \textbf{ACAC} & $\mathbf{G}_{\mathrm{neg}}$ & $\mathbf{G}_{\mathrm{pos}}$ \\
\midrule
10 & \ms{0.745}{0.007} & \ms{0.146}{0.002} & \ms{0.179}{0.001} & \ms{0.114}{0.002} \\
25 & \ms{0.740}{0.005} & \ms{0.120}{0.002} & \ms{0.160}{0.003} & \ms{0.080}{0.002} \\
50 & \ms{0.732}{0.004} & \ms{0.082}{0.003} & \ms{0.134}{0.004} & \ms{0.031}{0.001} \\
75 & \ms{0.726}{0.008} & \ms{0.061}{0.000} & \ms{0.124}{0.002} & \ms{-0.001}{0.002} \\
100 & \ms{0.704}{0.008} & \ms{0.049}{0.002} & \ms{0.148}{0.002} & \ms{-0.051}{0.003} \\
\midrule
ERM & \ms{0.739}{0.004} & \ms{0.186}{0.002} & \ms{0.221}{0.001} & \ms{0.151}{0.002} \\
\bottomrule
\end{tabular}}
\caption{ActorCorr sensitivity to $\lambda$, reported as mean $\pm$ standard deviation over three seeds. $G_y$ is the clean-to-anti-correlated accuracy gap for class $y$; negative values indicate a sign-inverted response.}
\label{tab:actorcorr-lambda}
\end{table}

Increasing $\lambda$ monotonically reduces ACAC from $0.146 \pm 0.002$ to $0.049 \pm 0.002$, but also lowers pooled OOD accuracy from $0.745 \pm 0.007$ to $0.704 \pm 0.008$. ActorCorr therefore reveals a clear accuracy--invariance trade-off: the penalty value that minimizes sensitivity to the targeted shortcut does not maximize aggregate predictive performance. Relative to the ERM reference, every IRM configuration reduces ACAC and both class-specific gaps, confirming that the invariance penalty consistently weakens reliance on the actor cue. However, only $\lambda\in\{10,25\}$ matches or improves upon ERM's OOD accuracy, whereas stronger penalties achieve greater invariance at the cost of predictive performance. The class-specific gaps further show that this trade-off is asymmetric. For the negative class, $G_{\mathrm{neg}}$ remains positive for all penalty values. For the positive class, the gap approaches zero at $\lambda=75$ and becomes negative at $\lambda=100$, with $G_{\mathrm{pos}}=-0.051 \pm 0.003$. The strongest penalty therefore induces a sign-inverted response for positive examples rather than simply removing sensitivity to the actor cue. This asymmetry may result from unequal associations of the two actor names in the frozen pretrained representation space. Consequently, a single global penalty can suppress the two class-specific shortcut responses at different rates.

\subsection{Sensitivity to Shortcut Prevalence}
\label{app:actorcorr-frequency}

The main ActorCorr experiment fixes the effective actor-injection frequency at
$5\%$. To assess how the prevalence of the shortcut changes the comparison,
we additionally generate two datasets in which $10\%$ or $25\%$ of reviews in
each training environment receive an actor substitution. We retain the same
actor--label purities, $\pi^{(e_1)}=1.0$ and $\pi^{(e_2)}=0.85$, the same
three evaluation conditions, and the same optimization protocol. Thus, this
intervention changes the prevalence of the actor cue without changing its
environment-specific agreement mechanism.

\begin{table}[t]
\centering
\scriptsize
\setlength{\tabcolsep}{1.5pt}
\renewcommand{\arraystretch}{1.08}
\resizebox{\columnwidth}{!}{
\begin{tabular}{@{}clcccc@{}}
\toprule
\textbf{Injection} & \textbf{Model} & \textbf{OOD} & \textbf{ACAC}
& $G_{\mathrm{neg}}$ & $G_{\mathrm{pos}}$ \\
\midrule
10\% & ERM       & \ms{0.732}{0.007} & \ms{0.160}{0.005} & \ms{0.203}{0.011} & \ms{0.118}{0.003} \\
     & IRM, 10   & \ms{0.741}{0.008} & \ms{0.124}{0.002} & \ms{0.174}{0.001} & \ms{0.075}{0.003} \\
     & IRM, 25   & \ms{0.739}{0.005} & \ms{0.100}{0.001} & \ms{0.138}{0.003} & \ms{0.062}{0.001} \\
     & IRM, 50   & \ms{0.734}{0.005} & \ms{0.071}{0.003} & \ms{0.115}{0.003} & \ms{0.027}{0.002} \\
     & IRM, 75   & \ms{0.725}{0.005} & \ms{0.055}{0.003} & \ms{0.115}{0.003} & \ms{-0.004}{0.004} \\
     & IRM, 100  & \ms{0.709}{0.005} & \ms{0.043}{0.000} & \ms{0.142}{0.002} & \ms{-0.055}{0.003} \\
\midrule
25\% & ERM       & \ms{0.717}{0.008} & \ms{0.291}{0.003} & \ms{0.324}{0.006} & \ms{0.258}{0.003} \\
     & IRM, 10   & \ms{0.724}{0.008} & \ms{0.227}{0.001} & \ms{0.256}{0.004} & \ms{0.198}{0.002} \\
     & IRM, 25   & \ms{0.728}{0.009} & \ms{0.170}{0.003} & \ms{0.202}{0.006} & \ms{0.137}{0.003} \\
     & IRM, 50   & \ms{0.728}{0.008} & \ms{0.115}{0.004} & \ms{0.164}{0.011} & \ms{0.065}{0.008} \\
     & IRM, 75   & \ms{0.716}{0.009} & \ms{0.098}{0.005} & \ms{0.162}{0.002} & \ms{0.033}{0.009} \\
     & IRM, 100  & \ms{0.697}{0.007} & \ms{0.071}{0.004} & \ms{0.176}{0.013} & \ms{-0.034}{0.005} \\
\bottomrule
\end{tabular}}
\caption{ActorCorr sensitivity to effective actor-injection frequency and
IRM penalty, reported as mean $\pm$ standard deviation over three seeds.
$G_y$ is the clean-to-anti-correlated accuracy gap for class $y$; negative
values indicate a sign-inverted response.}
\label{tab:actorcorr-frequency}
\end{table}

Table~\ref{tab:actorcorr-frequency} shows that the frequency manipulation
successfully controls shortcut attractiveness. Increasing the injection
frequency from $10\%$ to $25\%$ raises ERM's ACAC from
$0.160 \pm 0.005$ to $0.291 \pm 0.003$, while lowering its pooled OOD accuracy
from $0.732 \pm 0.007$ to $0.717 \pm 0.008$. At $25\%$, IRMv1 with
$\lambda=50$ is the best accuracy--invariance compromise: it improves OOD
accuracy to $0.728 \pm 0.008$ and reduces ACAC to $0.115 \pm 0.004$.
This corresponds to a $60.5\%$ reduction in signed average sensitivity
relative to ERM, while both class-specific gaps remain positive and smaller
than their ERM counterparts.

The lowest ACAC values are obtained with $\lambda=100$, but this setting also
reduces OOD accuracy and makes $G_{\mathrm{pos}}$ negative at both injection
frequencies. As in the fixed-frequency sweep, this behavior is evidence of a
sign-inverted response for the positive class rather than complete invariance.
At the weaker $10\%$ prevalence, the OOD optimum occurs at $\lambda=10$;
larger penalties trade accuracy for lower shortcut sensitivity. The frequency
sweep therefore confirms that the useful penalty scale depends on shortcut
prevalence, while preserving the same qualitative accuracy--invariance
trade-off observed in Table~\ref{tab:actorcorr-lambda}.

\end{document}